\documentclass[10pt,twocolumn,letterpaper]{article}

\usepackage[pagenumbers]{cvpr} 

\usepackage{placeins}  
\usepackage[labelformat=parens,labelsep=space]{subcaption}
\usepackage{colortbl}  

\definecolor{cvprblue}{rgb}{0.21,0.49,0.74}
\usepackage[pagebackref,breaklinks,colorlinks,allcolors=cvprblue]{hyperref}

\def\confName{CVPR}
\def\confYear{2027}
\newcommand{\lrtok}{\ensuremath{S_{\text{lo}}T_{\text{lo}}}}
\newcommand{\hrtok}{\ensuremath{S_{\text{hi}}T_{\text{hi}}}}

\newcommand{\Shi}{\ensuremath{S_{\text{hi}}}}

\title{MiSTR: Adaptively Mixing Spatiotemporal Resolution\\for Satellite Image Time Series}

\author{
Tim G. Zhou\textsuperscript{1,2} \quad
Anthony Fuller\textsuperscript{2} \quad
Avery Ma\textsuperscript{3} \quad
Geoff Pleiss\textsuperscript{1,2} \quad
Evan Shelhamer\textsuperscript{1,2} \\
\textsuperscript{1}University of British Columbia \quad
\textsuperscript{2}Vector Institute \quad
\textsuperscript{3}RBC Borealis \\
}

\begin{document}
\maketitle
\begin{abstract}
Earth Observation (EO) satellites continuously collect image time series, capturing changes and context at multiple temporal and spatial scales, from multi-year growth across whole forests, through seasonal cycles in individual crop fields, to sudden floods along a single river bank.
Current models encode these inputs at a single, fixed spatiotemporal resolution.
One size does not fit all tasks: high resolution may be inefficient, low resolution may be inaccurate, and in reality the level of detail needed varies across space and time.
We thus propose \underline{Mi}xed \underline{S}patio\underline{t}emporal \underline{R}esolution (MiSTR) for accurate and efficient adaptive computation of transformers for EO.
First, a full low-resolution pass provides coarse features and predictions, and selects where and when higher resolution is needed.
Then, a sparse high-resolution pass adds tokens at the selected spatiotemporal points to refine the predictions.
Across five datasets spanning crop-type segmentation (PASTIS, PLANTEUR), biomass (BioMassters) and yield regression (YieldSAT), and flood detection (GEOID-Flood), MiSTR achieves a better computation--performance tradeoff than fixed-resolution processing, and outperforms token merging, adaptive tokenization, and feature upsampling methods developed for RGB images and videos.
On long-series flood tracking, MiSTR outperforms fixed-resolution processing with $7\times$ less computation.
Code is available at \url{https://github.com/timgzhou/MiSTR}.
\end{abstract}    
\section{Introduction}
\label{sec:intro}

Earth observation (EO) satellites capture images with near-global spatial coverage and frequent temporal revisits.
Satellite image time series (SITS) support many applications: land-use mapping \citep{sykas2022sentinel}, agricultural parcel segmentation \citep{garnot2021panoptic,turkoglu2021crop}, crop yield estimation \citep{miranda_2026_yieldsat,Sani_2024_WACV}, and natural disaster mapping \citep{Chiriaco2026GEOIDFlood,gerard2023wildfirespreadts}.
These and many other real-world tasks require dense outputs \citep{marsocci2024pangaea}, yet each point in space and time may not require equal computation.
Some points in space may be harder to predict than others: in PASTIS crop-type segmentation \citep{garnot2021panoptic}, the boundary between fields may be more complex than the middle of a field.
Likewise, some points in time change more than others: in GEOID-Flood \citep{Chiriaco2026GEOIDFlood}, pixels during a flood are more dynamic than those of a permanent lake.
This presents an opportunity for efficiency by allocating more computation where and when it is needed, depending on the task and the instance itself.

The state-of-the-art methods on SITS benchmarks \citep{marsocci2024pangaea} are vision transformers (ViTs \citep{dosovitskiy2021an}) pre-trained on large-scale, multi-modal, multi-timestep satellite imagery.
By splitting inputs into grids of patches, ViTs allocate tokens equally over space and time.
The grid size is governed by the patch size, so smaller patches yield more tokens.
The computational cost of tokens grows superlinearly in their number, and SITS contain many tokens, easily 10K+ for one instance (Appendix~\ref{apx:token-counts}), so processing them can be prohibitively expensive.

In this work, we reduce the cost of processing SITS by adaptively re-tokenizing when and where it is most needed.
Our method, \textbf{MiSTR}, encodes a SITS with two forward passes that share a single set of weights.
A context pass encodes coarse temporal composites with large patches to produce coarse features and predictions.
From these, MiSTR selects where (spatial points of high uncertainty) and when (the seasons the decoder attends to most) finer resolution is needed.
A detail pass then re-encodes only the selected points with small patches and finer temporal resolution, refining the coarse predictions.
Crucially, we propose a one-way attention mechanism that lets the detail tokens read the cached context tokens without recomputing them, so the detail pass costs little.

Unlike token merging and pruning \citep{bolya2022token,choi2024vid}, MiSTR never processes all tokens at high resolution; unlike feature upsamplers \citep{wimmer2026anyup,Seo_2026_CVPR}, it goes back to the input to re-encode the selected spatiotemporal points \citep{tseng2025galileo,herzog2026olmoearth}.
MiSTR adapts the multi-pass refinement of PointRend \citep{kirillov2020pointrend} to flexible-patch EO transformers, and does so without requiring any additional model parameters.

We evaluate MiSTR on five dense EO benchmarks: crop-type segmentation (PASTIS \citep{garnot2021panoptic}, PLANTEUR \citep{planteur2026}), above-ground biomass regression (BioMassters \citep{nascetti2023biomassters}), crop yield regression (YieldSAT \citep{miranda_2026_yieldsat}), and flood mapping (a custom-built SITS benchmark from GEOID-Flood \citep{Chiriaco2026GEOIDFlood}).
Across these datasets, MiSTR (initialized with OlmoEarth \cite{herzog2026olmoearth}) achieves a better computation--performance tradeoff than fixed-resolution processing, matching the accuracy of the best fixed-resolution OlmoEarth across computational budgets with 1.3--4.6$\times$ fewer GMACs, and achieving better absolute performance at matched computation.
On GEOID-Flood, MiSTR tracks floods across a 32-timestep series, matching fixed-resolution OlmoEarth with $7\times$ fewer GMACs and surpassing it at the same cost.
We further show that, at matched computation, MiSTR outperforms state-of-the-art token merging (vid-TLDR~\citep{choi2024vid}), adaptive tokenization (APT~\citep{choudhury2025accelerating}) and feature upsampling (AnyUp~\citep{wimmer2026anyup}, Upsample Anything~\citep{Seo_2026_CVPR}) methods developed for RGB images and videos on SITS benchmarks.

\begin{figure*}[t]
    \centering
    \includegraphics[width=1\linewidth]{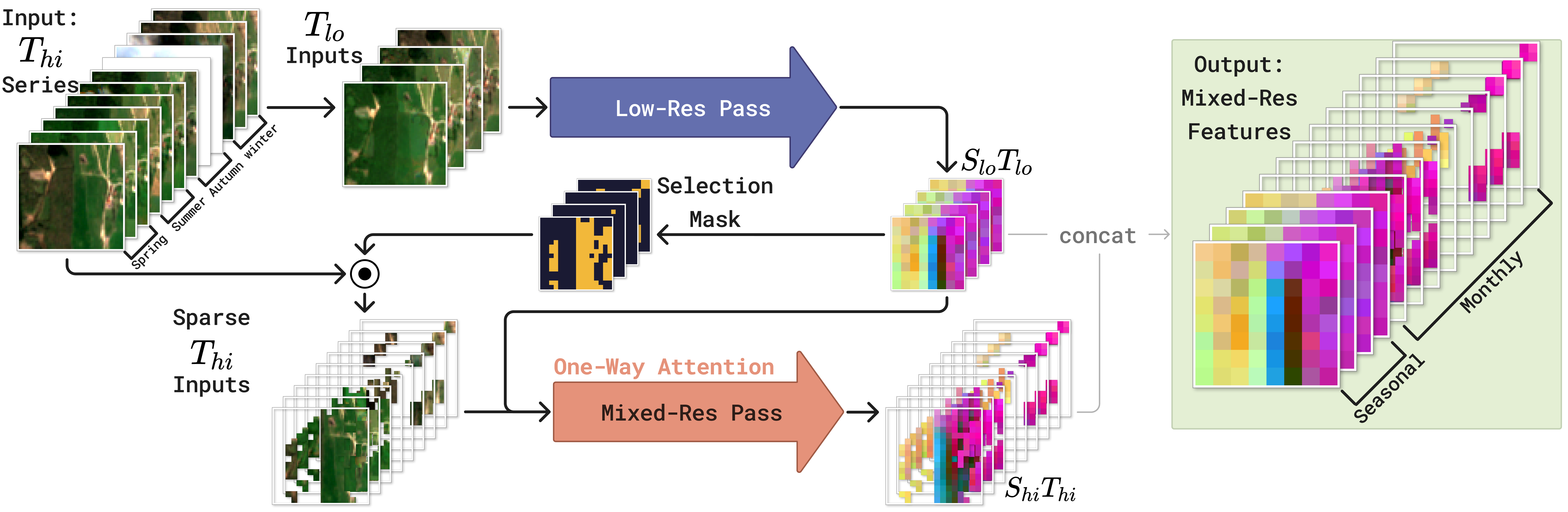}
    \caption{\textbf{MiSTR transformers process remote sensing time series in two passes with a single set of weights.}
    A first pass over $\text{T}_\text{lo}$ composites at low spatial resolution extracts dense \lrtok{} features and selects which spatiotemporal tokens to re-encode. 
    A second, mixed-resolution pass re-encodes the selected sparse \hrtok{} tokens from the monthly series through one-way attention. 
    The output is a mixed-resolution feature map: dense seasonal features plus sparse monthly features.}
    \label{fig:method}
\end{figure*}
\section{Background}
\label{sec:existing_work}

\textbf{Spatiotemporal resolution in remote sensing.}
Satellite image time series (SITS) can vary in their spatial and temporal resolutions.
Freely available sensors range from 10\,m (Sentinel-1 and -2) to 30\,m (Landsat) per pixel, and revisit a location every 5 (Sentinel-2), 6--12 (Sentinel-1) or 8 days (Landsat 8 and 9).
A vision transformer (ViT) \citep{dosovitskiy2021an} then sets two more resolutions on top of the sensor's: the patch size sets the spatial resolution of its tokens, and the temporal sampling (every acquisition, or monthly or seasonal composites) sets their temporal resolution.
The number of tokens grows as either resolution becomes finer, so no single spatiotemporal resolution is both cheap and accurate for every task.
Spatially, most pre-trained EO ViTs fix the patch size at pre-training \citep{cong2022satmae,jakubik2023foundation,szwarcman2025prithvi,fuller2023croma}; some flexible-patch models defer the choice to fine-tuning \citep{tseng2025galileo,herzog2026olmoearth}.
Temporally, series longer than the pre-training length are split into independent chunks \citep{cong2022satmae}, aggregated by an extra temporal encoder \citep{szwarcman2025prithvi}, or averaged into monthly composites \citep{herzog2026olmoearth}, each fixing one temporal resolution for the whole input.
In all cases, the chosen spatiotemporal resolution applies to every location and timestep of every input, so practitioners must commit to one for their problem.
MiSTR removes this commitment: by adaptively choosing the resolution per point in space and time for each input, MiSTR uses fine resolution only where and when it is needed.

\textbf{OlmoEarth.}
Our main results report MiSTR OlmoEarth models, which are built on the OlmoEarth foundation model~\citep{herzog2026olmoearth}.
OlmoEarth is a simple multimodal ViT that takes Sentinel-2, Sentinel-1 and Landsat time series as input.
Each modality's bands are split into a few band groups, and each band group at each timestep and patch location becomes one token.
Patches of any size are tokenized with a single set of embedding weights per band group, by interpolating each input patch to the kernel size, so coarse and fine patches are encoded into a shared token space.
The number of tokens therefore depends on the modalities, the spatial extent and length of the series, and the chosen patch size.
Each token's embedding is summed with four encodings, each occupying a quarter of the embedding dimensions: a learned sensor encoding (one per band group), and sinusoidal encodings of its time index (its position in the series), its month, and its 2D spatial position.
All tokens from all timesteps and modalities then pass through a standard transformer encoder with full self-attention.

\section{Method: \underline{Mi}xing \underline{S}patio\underline{t}emporal \underline{R}esolution}
\label{sec:method}

MiSTR aims to reduce the computation of models for SITS while keeping high spatiotemporal resolution details where they matter.
We define a token's spatiotemporal resolution by its stride: its spatial stride is the distance between neighboring tokens, and its temporal stride is the interval between consecutive tokens, from a season down to a single acquisition.
SITS can then be tokenized at any spatiotemporal resolution between these two extremes:
\begin{itemize}
\item \emph{Coarse everywhere:} large spatial and temporal strides are efficient, but miss fine structures and short-lived changes.
\item \emph{Fine everywhere:} small spatial and temporal strides are precise, but require significantly more computation. 
\end{itemize}
MiSTR sits between the two: it first encodes the SITS with low spatiotemporal resolution (\lrtok{} tokens), and adaptively re-encodes it at higher spatiotemporal resolution (\hrtok{} tokens) only where and when needed.
MiSTR does this by processing each instance in two passes that share one set of weights (\cref{fig:method}).
\begin{enumerate}
\item The first pass encodes the series as \lrtok{} tokens and makes a coarse prediction.
A selector highlights where and when finer detail is needed based on spatial uncertainty and the temporal attention of the coarse prediction.
\item The second pass re-encodes the selected spatiotemporal points as \hrtok{} tokens, which attend to other \hrtok{} tokens and the cached \lrtok{} tokens from the first pass to refine the coarse prediction.
\end{enumerate}
Carrying out this MiSTR design entails four design choices, which we go over in the following subsections.
\begin{itemize}
\item \textbf{Section~\ref{subsec:tokenization}: Patches of different sizes} are tokenized with shared weights, which adds no parameters and keeps \lrtok{} and \hrtok{} tokens in a common space.
\item \textbf{Section~\ref{subsec:selection}: Selecting spatiotemporal points} ensures that extra computation goes only to the places and times that need it most.
\item \textbf{Section~\ref{subsec:one-way-attention}: One-way attention}, where \hrtok{} tokens attend to \lrtok{} tokens but not the reverse in the second pass, avoids recomputing the first pass and gives \hrtok{} tokens context.
\item \textbf{Section~\ref{subsec:decoder}: A decoder for tokens of different spatiotemporal resolution} produces a prediction at every pixel, however many tokens cover it.
\end{itemize}

\subsection{Tokenizing Patches of Different Sizes}
\label{subsec:tokenization}
We tokenize \lrtok{} and \hrtok{} patches with OlmoEarth's shared patch embedding layer, keeping its division of sensor band groups, and sum each token's embedding with OlmoEarth's four encodings (\cref{sec:existing_work}).
The \emph{sensor encoding} is unchanged: it is learned per band group and shared by both token types.
Because \lrtok{} and \hrtok{} tokens cover periods of different lengths and patches of different sizes, we define the other three encodings so that tokens covering the same place and time period receive coherent encodings:
\begin{itemize}
\item \emph{Time-index encoding:} each \lrtok{} token's time index is its position in the series, and each \hrtok{} token takes the time index of the \lrtok{} token whose time span contains it.
\item \emph{Month encoding:} each token uses its actual month, or the average calendar month if it covers several months. In our main experiments, this is the exact month for \hrtok{} tokens and the average month for \lrtok{} tokens (\cref{sec:results}).
\item \emph{Spatial position encoding:} we encode the position of each patch's center pixel rather than its top-left corner.
Because tokens have different patch sizes, this places a large patch at the mean position of the small patches that tile it.
\end{itemize}

\subsection{Selecting Spatiotemporal Points}
\label{subsec:selection}
MiSTR must select the spatiotemporal points whose re-encoding as \hrtok{} tokens gives the largest improvement for the least additional computation.
Intuitively, selecting in space and in time calls for different signals.
Spatially, we want to re-encode the points where the coarse prediction is likely to be wrong, which its uncertainty indicates.
Temporally, we want to zoom in on the observations that carry the most information for that location, which the decoder's temporal attention reveals.
Both signals come from the \lrtok{} pass for free, requiring no additional parameters.

\textbf{Selecting where.}
To select spatially uncertain regions, we use task-dependent heuristics for predictive uncertainty.
For classification, we measure uncertainty with the Shannon entropy of the coarse prediction.
For regression, we use the variance of the per-timestep predictions, which is high where the timesteps disagree.
We average this uncertainty over time for each candidate \Shi{} point and select with a shared threshold set to the $(1 - q)$ quantile of the uncertainty over the validation set, where $q$ is a user-defined selection percentage of all possible \hrtok{} points.
A fixed threshold, rather than a fixed number of points per sample, allows MiSTR to spend more of the budget on difficult samples and less on easy ones.

\textbf{Selecting when.}
For each selected \Shi{} point, we take the \lrtok{} token with the highest decoder attention score (\cref{subsec:decoder}) among all spatially overlapping \lrtok{} tokens, and re-encode all \hrtok{} points within its time span in the second pass.

Despite their simplicity, these heuristic signals are highly effective, and re-encoding their selections consistently improves downstream performance compared to naive selections (Appendix~\ref{apx:selection}).

\subsection{One-Way Attention from Coarse to Fine}
\label{subsec:one-way-attention}
The first pass used for \hrtok{} point selection already encodes every \lrtok{} token through the full transformer.
Encoding all tokens jointly in a second pass would repeat this work, so we develop a one-way attention mechanism, which lets information flow from \lrtok{} to \hrtok{} but not the reverse.
\begin{figure}[h]
    \centering
    \includegraphics[width=0.9\linewidth]{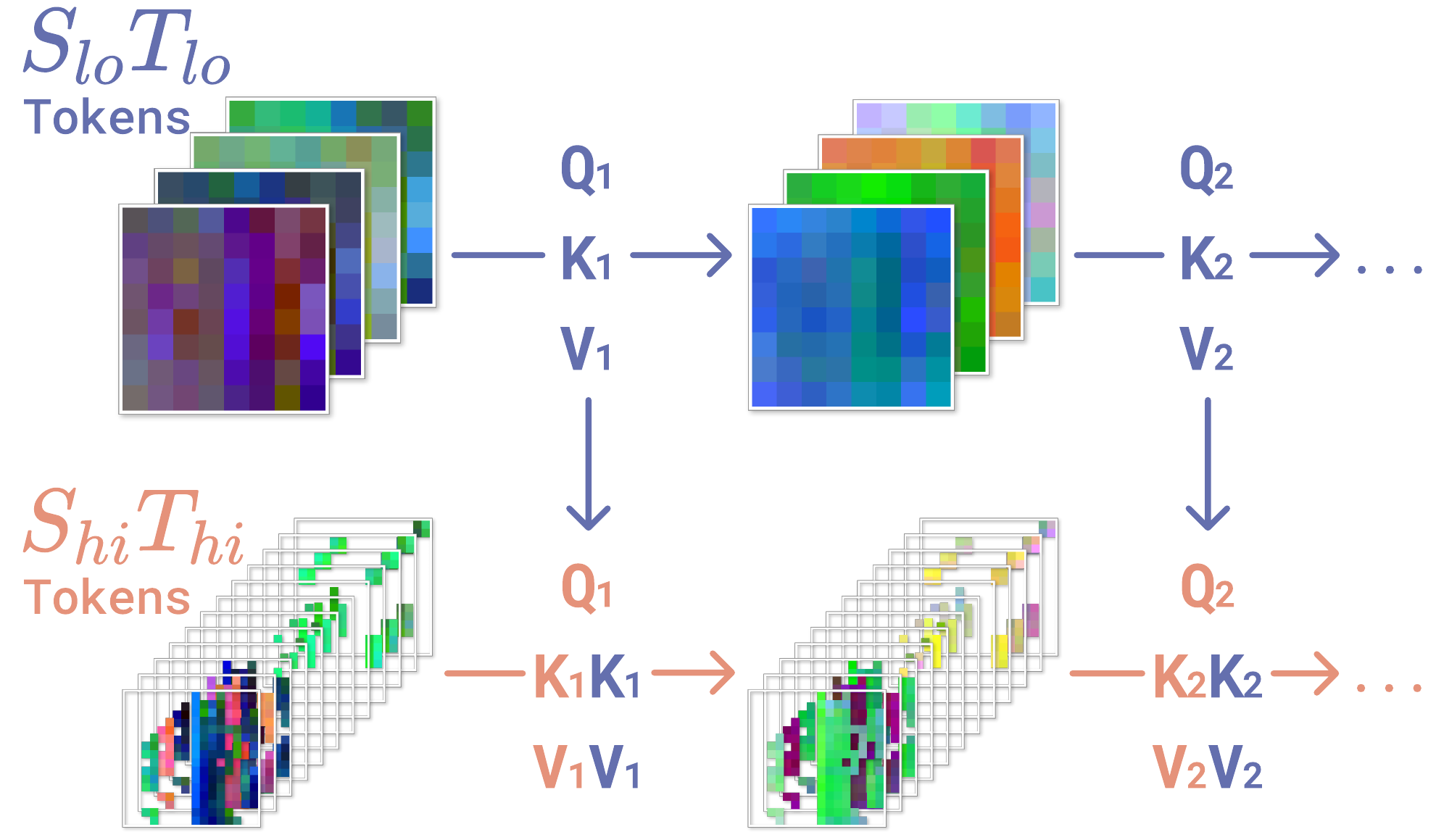}
    \caption{\textbf{One-way attention encodes tokens of mixed spatiotemporal resolution while saving computation.} \lrtok{} tokens (left) and selected \hrtok{} tokens (right) pass through the same transformer layers with shared weights. \lrtok{} tokens are calculated in the first pass, attending only to each other; \hrtok{} tokens attend to each other and the \lrtok{} tokens from the matching layer, which are cached from the first pass.}
    \label{fig:one-way-attention}
\end{figure}
Figure~\ref{fig:one-way-attention} illustrates this one-way attention mechanism.
At each layer, the selected \hrtok{} tokens act as queries over the sample's \lrtok{} tokens from that layer and over all \hrtok{} tokens, using the shared encoder weights.
\lrtok{} tokens never attend to \hrtok{} tokens, so they are unchanged by the second pass and can be cached from the first.

With $N_\text{lo}$ \lrtok{} tokens and $N_\text{hi}$ \hrtok{} tokens, jointly re-encoding them through standard self-attention in the second pass costs $O((N_\text{lo} + N_\text{hi})^2)$ attention operations per layer, whereas one-way attention costs only $O(N_\text{hi} (N_\text{lo} + N_\text{hi}))$, because it needs no queries from \lrtok{} tokens; the per-token projections and MLPs likewise run only on the $N_\text{hi}$ \hrtok{} tokens, saving an additional $O(N_\text{lo})$ MLP passes.
Allowing \hrtok{} tokens to read each other, rather than only the \lrtok{} tokens, lets selected regions exchange fine detail across space and time.

Appendix~\ref{apx:one-way-attention-ablation} compares the one-way attention mechanism to full self-attention and cross-attention. That ablation verifies empirically that one-way attention provides substantial computational savings while effectively providing the necessary context for \hrtok{} tokens.

\subsection{Decoding Mixed Tokens to Output Resolution}
\label{subsec:decoder}
MiSTR produces features at mixed spatiotemporal resolution: at any pixel, there can be multiple \lrtok{} tokens, potentially overlapping with multiple \hrtok{} tokens.
To make per-pixel predictions, we decode every token into its own predictions and optionally pool per-pixel predictions over time if the task requires one prediction map for all timesteps.
A linear head $h$, shared across all tokens, maps the feature $f_i$ of token $i$ to its prediction map $\hat{y}_i = h(f_i)$, the size of an \lrtok{} patch; for an \hrtok{} feature token, we average-pool this map to its own patch size, so that a token of patch size $p$ makes $p \times p$ pixel predictions.
These outputs are class logits for segmentation and point estimates for regression; the decoder treats both identically.
Let $\mathcal{T}[u, v]$ be the set of tokens whose patch contains pixel $(u, v)$.
The prediction at pixel $(u, v)$ is an attention-weighted mean of these tokens' outputs at that pixel:
\begin{equation}
\begin{aligned}
    \hat{y}[u, v] &= \sum_{i \in \mathcal{T}[u, v]} w_i[u, v] \, \hat{y}_i[u, v],\\
    w_i[u, v] &= \operatorname{softmax}_{i \in \mathcal{T}[u, v]} \, s(f_i),
\end{aligned}
\end{equation}
where $s(f_i) = \mathbf{u}^\top \mathrm{LN}(f_i)$ scores each token against a learned query $\mathbf{u}$, so the decoder is attention pooling over the tokens covering each pixel; $\mathbf{u}$ is initialized to zero, so that the decoder starts as a mean over time.

This decoder has three useful properties.
It accepts any number of tokens per pixel at any patch size, so it needs no fixed temporal grid.
Pixels with no selected tokens receive exactly the prediction of the \lrtok{} tokens due to the one-way attention mechanism.
Finally, the score $s$ provides the attention signal for the temporal selector, so time selection needs no additional training.
We use the same decoder for the fixed-resolution baselines, so that differences in \cref{sec:results} reflect the tokenization rather than the decoder.

\begin{figure*}[t]
    \centering
    \includegraphics[width=1\linewidth]{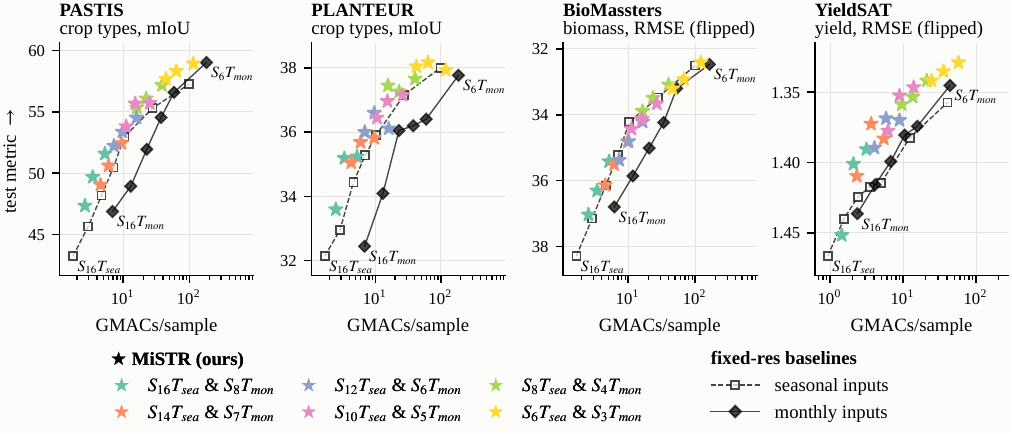}
    \caption{\textbf{MiSTR is cheaper and better across dense SITS tasks.} MiSTR OlmoEarth models over a simple sweep of resolutions are consistently above the performance-computation Pareto frontier of fully fine-tuned fixed-spatiotemporal-resolution models on PASTIS, PLANTEUR and YieldSAT, and tie the best fixed-resolution tradeoff on BioMassters.}
    \label{fig:result}
\end{figure*}

\section{Experiments}
\label{sec:results}

\subsection{Experimental Setup}
\label{subsec:setup}

\textbf{Datasets.}
We evaluate on four dense prediction tasks in which a model sees a whole series but only needs to make one per-pixel prediction.
\begin{itemize}
\item \textbf{PASTIS} \citep{garnot2021panoptic} contains 2,433 patches of $128\times128$ pixels over France, with semantic labels for 18 crop types.
We use its Sentinel-2 and Sentinel-1 (ascending) time series, which span September 2018 to November 2019.
Different crop types often look alike in certain seasons, so segmentation models benefit from access to temporal data for differentiating how fields develop over the season.
\item \textbf{PLANTEUR} \citep{planteur2026} follows the PASTIS format and covers six tropical regions in the French overseas territories, with 7 crop classes.
We use Sentinel-2 and Sentinel-1.
\item \textbf{BioMassters} \citep{nascetti2023biomassters} is a pixel-wise regression benchmark over Finnish boreal forest with above-ground biomass as targets.
Each sample has one Sentinel-1 and one Sentinel-2 image per month for a year, making this a multimodal and multi-temporal dataset.
\item \textbf{YieldSAT} \citep{miranda_2026_yieldsat} is a recent, first-of-its-kind dataset for 10\,m resolution crop yield prediction. YieldSAT contains 2,173 fields across 4 countries, 4 crop types and 9 years.
We use only its Sentinel-2 L2A time series.
\end{itemize}
We composite each series into calendar months and group every three months into a season; months without an acquisition produce no tokens, for MiSTR and fixed-resolution models alike (Appendix~\ref{apx:datasets}).
\Cref{sec:flood} adds a fifth task, flood tracking on GEOID-Flood, where the model predicts a map at every acquisition.

\textbf{Fixed-resolution OlmoEarth.}
Our main baseline is OlmoEarth tokenized at fixed spatiotemporal resolution for the whole series: seasonal composites with patch sizes from 16 down to 4, or monthly composites with patch sizes from 16 down to 6.
It uses the same decoder as MiSTR (\cref{subsec:decoder}), so the two differ only in their tokens.

\textbf{MiSTR configurations.}
We use seasonal composites as $T_\text{coarse}$ and monthly composites as $T_\text{fine}$.
We pair each coarse patch size $S_\text{coarse} \in \{16, 14, 12, 10, 8, 6\}$ with $S_\text{fine} = S_\text{coarse}/2$, and each candidate point is one \hrtok{} patch.
We train each pair at three selection budgets, $q \in \{1/8, 1/4, 1/2\}$, the share of all candidate points selected for re-encoding.
Each selected point is re-encoded at every month of its selected season.
Appendix~\ref{apx:separate} motivates seasonal composites for the context pass and monthly composites for the detail pass, and Appendix~\ref{apx:flexible-time-budget} varies how many seasons each selected point re-encodes.

\textbf{Training and evaluation.}
All models start from OlmoEarth-Tiny \citep{herzog2026olmoearth} and are fine-tuned end to end on $64\times64$-pixel SITS with OlmoEarth's fine-tuning recipe: the encoder is frozen for the first 20\% of epochs and then trained at a tenth of the learning rate.
Appendix~\ref{apx:alt-rsfm-init} shows that MiSTR also applies to other foundation models: the non-temporal ViT TerraMind \citep{jakubik2025terramind} and the pixel time-series transformer Tessera (v2) \citep{feng2026tessera}.
We use AdamW with a learning rate of $10^{-3}$, reduced on plateaus of the validation metric, and train for 100 epochs (20 on BioMassters), picking the epoch with the best validation score.
MiSTR models are trained with a warm-up selector that selects all seasons for the high-loss patches before switching to entropy and attention spatiotemporal selections.
We report mIoU for segmentation and RMSE for regression on the test split, averaged over three seeds (Appendix~\ref{apx:seeds}).
Because MiSTR's selection adapts to each sample, harder samples receive more computation, so we measure MiSTR's compute cost (GMACs per SITS) directly on the test set.

\begin{figure*}[t]
    \centering
    \includegraphics[width=\textwidth]{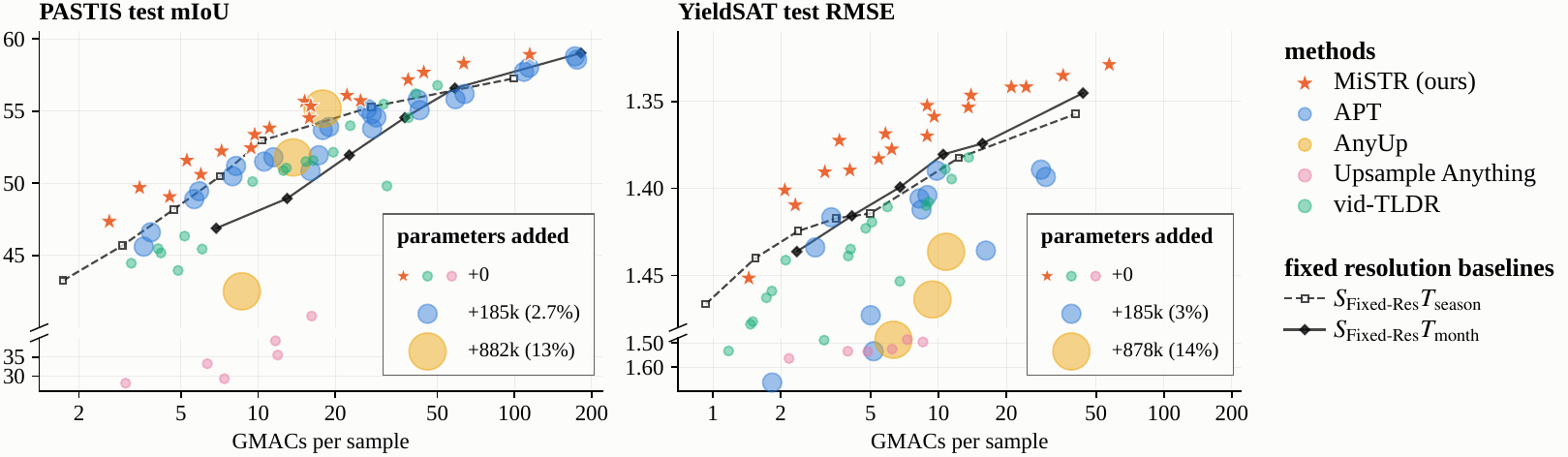}
    \caption{
    \textbf{MiSTR is parameter- and computation-efficient.}
    On PASTIS (left, test mIoU) and YieldSAT (right, test RMSE, lower is better and plotted upward), MiSTR adds no parameters and gives the best accuracy–compute tradeoff against adaptive tokenization (APT), training-free video token merging (vid-TLDR) and feature upsampling (AnyUp, Upsample Anything).
    Marker area is the number of parameters added to the fixed-resolution model; points past the axis break are drawn in a compressed strip.
    }
    \label{fig:baselines}
\end{figure*}

\textbf{Efficient ViT baselines.}
On PASTIS and YieldSAT, we also compare MiSTR with three families of methods that reduce the cost of dense prediction with ViTs, all developed for RGB images or videos.
Upsamplers start from a coarse fixed-resolution model, vid-TLDR from a fine one, and APT is fine-tuned with its mixed patch sizes; we sweep each method's resolution and merging or splitting rate to trace its accuracy against GMACs.
\emph{Feature upsamplers} upsample ViT features or predictions from patch resolution to (up to) pixel resolution, guided by the input, so fine details are weighted interpolations of coarse features or predictions rather than encoded from raw inputs.
\textbf{AnyUp} \citep{wimmer2026anyup} is a pre-trained upsampler that works with any backbone's features, guided by an RGB image (here, the Sentinel-2 RGB bands), and Upsample Anything (\textbf{UPA}) \citep{Seo_2026_CVPR} fits anisotropic Gaussian kernels to each image at test time and can upsample either features or predictions.
\emph{Token merging} starts from fine tokens and merges similar ones as the network gets deeper, whereas MiSTR starts from coarse tokens and adds fine ones where they are needed.
\textbf{vid-TLDR} \citep{choi2024vid} merges the tokens of a video without training, using attention to find salient regions and merging background tokens first.
\emph{Adaptive tokenization} splits the input into patches of different sizes before the encoder, so, as in MiSTR, each region gets its own spatial resolution, but the split is chosen from the input alone rather than from the model's own uncertainty, and the temporal resolution stays fixed.
\textbf{APT} \citep{choudhury2025accelerating} chooses patch sizes by pixel entropy and embeds large patches with a small learned network, so that they can be processed alongside small ones.
We discuss these baselines and other related methods in adaptive computation in more detail, and compare their mechanisms to MiSTR's, in \cref{sec:related_work}.

\subsection{Results: Many Inputs, One Output}
\label{subsec:main-results}
\Cref{fig:result} compares MiSTR OlmoEarth with fixed-resolution OlmoEarth, both fully fine-tuned, across spatiotemporal resolutions.
On PASTIS, PLANTEUR and YieldSAT, MiSTR lies above the fixed-resolution curves: at the same compute it is more accurate, and at the same accuracy it needs less computation.
On BioMassters it matches the best fixed-resolution tradeoff, the seasonal curve, and lies well above the monthly one.
Across the four datasets, MiSTR matches the accuracy of the best fixed-resolution OlmoEarth at each compute level with 1.3--4.6$\times$ fewer GMACs, and achieves better absolute performance at matched computation (Appendix \ref{apx:matched-compute}).
Under the same maximum computational budget, MiSTR's most expensive configuration, $S_6T_\text{sea}$ with $S_3T_\text{mon}$, re-encodes selected points as $3\times3$-pixel monthly tokens, a finer spatial resolution than any fixed-resolution model comparable in cost.
As a result, it matches or beats the most accurate fixed-resolution model on every dataset at lower cost ($S_6T_\text{mon}$, or $S_4T_\text{sea}$ on PLANTEUR): it is better on PLANTEUR at 43\% of its GMACs, on YieldSAT at 82\% and on BioMassters at 74\%, and comparable (within 0.1 mIoU) on PASTIS at 63\%.
Despite its simplicity and adding no new parameters, MiSTR achieves a better or equal accuracy--compute tradeoff across dense SITS tasks by allocating \hrtok{} tokens strategically across space and time through two passes. 

\textbf{MiSTR Spatiotemporal Selection Analysis.}
MiSTR selects spatial tokens using predictive entropy, and selects temporal tokens using the decoder's temporal attention. 
This can be interpreted in two steps: MiSTR first selects the locations where the overall prediction is most uncertain and, for these regions, zooms into their most informative season.
Appendix~\ref{apx:selection-samples} shows what these selectors select in practice.
Appendix~\ref{apx:selection} compares each signal against random and grid picks: entropy is the best for where (\cref{tab:selector}) and decoder attention for when (\cref{tab:time-selector}).

\textbf{Comparison to Baselines.}
\label{subsec:baselines}
\Cref{fig:baselines} compares MiSTR with the efficient ViT baselines of \cref{subsec:setup} on PASTIS and YieldSAT.
MiSTR gives the best accuracy at every computational budget while adding no parameters.
While APT (adaptive tokenization) and vid-TLDR (token merging) can occasionally outperform fixed-resolution OlmoEarth models, they generally trail $T_\text{sea}$ models in low-computation regimes and $T_\text{mon}$ models in high-computation regimes. 
AnyUp (pre-trained feature upsampler) is competitive in one PASTIS configuration and trails the fixed-resolution models elsewhere, while adding the most new parameters of the methods we benchmark. 
UPA (test-time feature upsampler) does not perform well, likely because of noise common in the RGB bands of Sentinel-2 images, such as cloud cover.
These results show that MiSTR outperforms state-of-the-art methods from RGB image and video processing when tested on SITS.
Appendix~\ref{apx:throughput} confirms these savings in measured GPU latency.

\subsection{Results: Many Inputs, Many Outputs}
\label{sec:flood}

One distinctive use of long satellite image time series is change detection.
Beyond the four benchmarks for per-pixel classification and regression (many inputs, one output), we show that MiSTR is well suited to this task: it processes long series cheaply, which matters for change detection in realistic settings.

\textbf{Setup.}
GEOID-Flood \citep{Chiriaco2026GEOIDFlood} provides a 3-class dry land / permanent water / flood segmentation target for pre- and post-flood Sentinel-1 images.
In practice, however, neither the time nor the place of a flood is known when it first happens, so a monitoring system must detect it without being told the event date.
A flood also rarely appears and disappears within one acquisition: it can worsen, linger or recede over weeks, so it must be tracked over many acquisitions rather than compared once before and after.
We therefore turn GEOID-Flood into a tracking task: for each event, we additionally fetch a dense series of up to 49 Sentinel-1 timesteps from the same orbit from the Microsoft Planetary Computer. The model sees a window of 32 consecutive timesteps with the labeled date at an unknown position, and must segment dry land, permanent water and flood at every timestep.
The model attends causally, so its prediction for a timestep uses only that timestep and earlier ones, as a real-time monitoring system would.
Compute cost is measured as the total computation up to the prediction for the labeled date.
Because only one date per series is labeled, we train with CutMix (Appendix~\ref{apx:geoid-flood}) and evaluate 3-class mIoU and flood boundary IoU (bIoU) on the labeled date.
Each acquisition is its own timestep, so MiSTR here mixes only spatial resolution and adaptively allocates its \hrtok{} budget across timesteps: it pairs a coarse patch size of 128 or 64 with a fine patch size of half or a quarter of it, and one model per pair is trained over a range of budgets.
Backbone, tile size and training details are in Appendix~\ref{apx:geoid-flood}.

\begin{figure}
    \centering
    \includegraphics[width=1\linewidth]{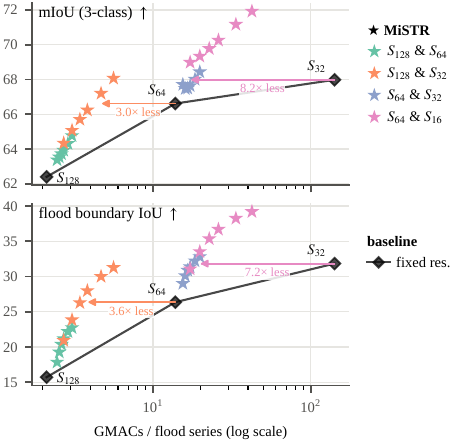}
    \caption{\textbf{MiSTR reaches the fixed-resolution scores at a fraction of the compute, then keeps improving.} 
    Flood tracking on GEOID-Flood series: 3-class mIoU and flood bIoU on the labeled flood date against the compute needed to predict it.
    Models attend causally in time, so this is the cost of the timesteps up to and including the labeled date; scores and costs are averaged over series with the labeled date at timestep 12, 16, 20 and 24 of 32.
    Arrows mark the compute saving at matched performance.}
    \label{fig:geoid}
\end{figure}
\begin{figure*}
    \centering
    \includegraphics[width=1\linewidth]{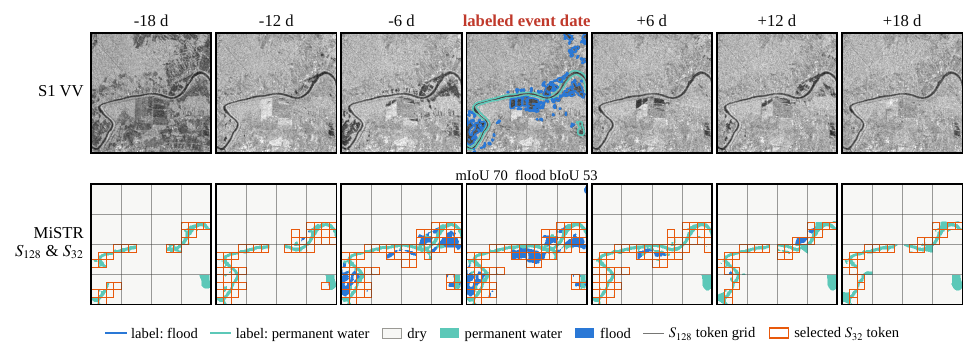}
    \caption{
    \textbf{Spatiotemporal change detection on GEOID-Flood$^*$ with higher accuracy and efficiency.} 
    MiSTR ($S_{128}$ \& $S_{32}$) improves on fixed-resolution $S_{32}$ with better accuracy (+2 at 70 mIoU / +3 at 53 flood bIoU) with fewer tokens (18.1\% at 1.5k tokens).
    Its fine tokens ($S_{32}$, orange) concentrate where and when the flood appears and recedes, while its coarse tokens ($S_{128}$) cover the stable river and land. Each timestep is a $512\times512$-pixel Sentinel-1 tile.
    $^*$We extend each before/after pair to a 32-timestep series to simulate real-world monitoring, where the event date is unknown; shown is a 7-timestep excerpt around the event.
    }
    \label{fig:geoid-sample}
\end{figure*}

\textbf{Results.}
\Cref{fig:geoid} shows that MiSTR's advantage is significant in this setting, because changes are localized in space and time.
MiSTR mixing $S_{64}$ and $S_{32}$ tokens matches fixed $S_{32}$ (68.4 vs.\ 68.0 mIoU, 32.8 vs.\ 31.8 flood bIoU) with about $7\times$ less compute.
At the same compute, mixing $S_{64}$ with $S_{16}$ already surpasses fixed $S_{32}$ on both metrics, and reaches 71.9 mIoU and 39.2 flood bIoU at less than a third of its compute: $S_{16}$ tokens over the whole series would be too expensive for a fixed-resolution model, but MiSTR can afford them where and when the flood appears.
\Cref{fig:geoid-sample} qualitatively shows MiSTR at work: it allocates its \hrtok{} tokens to where and when change is suspected, and tracks the flood from onset to recession.

\section{Related Work}
\label{sec:related_work}
Balancing accuracy and efficiency by adaptive computation is well established for vision \cite{han2021dynamic}, but remote sensing data is different \cite{rolf2024position}, so we bring spatiotemporal adaptive computation into orbit for satellite image time series.

\textbf{Coarse-to-fine computation} goes from low to high by adding points.
PointRend \citep{kirillov2020pointrend} iteratively adds finer points to coarse spatial predictions, guided by uncertainty, by interpolation of convolutional features and refinement by an additional model. 
Glance-and-focus methods \citep{huang2022glance} iteratively pick and process spatial patches, chosen by a policy learned with reinforcement learning, from coarse to fine as an efficient cascade of global and local models for classification.
Inspired by their multiple resolutions over multiple passes,
MiSTR adds points in space and time, guided by uncertainty and attention, and \emph{re}-encodes the input itself at higher resolution with the same model.

Feature interpolation or upsampling can convert low resolution to high resolution, even using the input as guidance, by signal processing (with joint bilateral filtering), learning (end-to-end), or test-time optimization \citep{fu2024featup,couairon2026jafar,ranzinger2025featsharp,wimmer2026anyup,Seo_2026_CVPR}.
These methods estimate high-resolution representations from low-resolution information directly, never re-encode the input with the backbone, and spend the same computation on every pixel. 
MiSTR, on the other hand, uses the low-resolution information to select the spatiotemporal points worth re-encoding at high resolution.

\textbf{Fine-to-coarse computation} goes from high to low by removing points.
Token reduction drops tokens from layer to layer to accelerate attention:
DynamicViT \citep{rao2021dynamicvit} learns to score token importance and progressively drops the lowest, EViT \citep{liang2022not} keeps high-attention tokens and merges low-attention tokens (judged by patch/CLS attention), 
and ToMe \citep{bolya2022token} merges similar tokens by bipartite matching.
vid-TLDR \citep{choi2024vid} merges tokens in videos by measuring attention in space and time, which brings it closer to our setting, so we extend it from global to local prediction by mapping merged tokens back to their original coordinates.
These methods reduce tokens, with an exhaustive pass for fine tokens, while MiSTR adds tokens, with a selective pass for fine tokens only where and when needed, conditioned on fewer coarse tokens.

\textbf{Adaptive tokenization.}
FlexiViT \citep{beyer2023flexivit} trains one ViT for many patch sizes, which OlmoEarth and MiSTR build on to tokenize coarse and fine patches with shared weights.
Mixed-resolution tokenization \citep{ronen2023vision,choudhury2025accelerating} splits an image into patches/tokens of variable size to balance accuracy (more, smaller tokens) with efficiency (fewer, bigger tokens).
The split is chosen from the raw input by frequency or entropy, and can require additional parameters. 
LookWhere \citep{fuller2026lookwhere} and its video extension LookWhen \citep{salamatian2026lookwhen} pair a shallow selector, which scores the tokens of a downscaled input, with a deep extractor, which encodes only the top-scoring high-resolution tokens in space (LookWhere) or in space and time (LookWhen).
The selector and extractor are separate networks that require costly joint pre-training by distillation from large self-supervised teachers before fine-tuning.
MiSTR instead runs both passes with the same pre-trained backbone and selects from the uncertainty and task-specific attention of its own coarse prediction for the task, rather than from task-agnostic pre-trained scores.
It therefore adds no parameters, requires no pre-training stage, and applies directly to existing remote sensing foundation models.

\section{Conclusion}
\label{sec:conclusion}
Existing models process a satellite image time series at one spatiotemporal resolution, unconditionally computing every point in space and time even where and when it is unnecessary.
MiSTR encodes the whole series coarsely and re-encodes it finely only where and when its own coarse prediction calls for it, with shared weights, one-way attention and a decoder for tokens of mixed resolution, adding no parameters yet achieving a better computation--performance tradeoff across benchmarks.
In flood tracking, where changes are localized in space and time, the benefit is largest: MiSTR matches fixed-resolution models with $7\times$ less compute and surpasses them at the same cost.

\textbf{Limitations.}
MiSTR's selection signals are spatial uncertainty and temporal attention of the coarse prediction.
They are cheap to compute, require no new parameters, and are effective despite their simplicity.
Nonetheless, better selectors could widen MiSTR's advantage; they could encode domain expertise as other heuristics, or be learned end-to-end with new parameters.
Similarly, for simplicity we fix the fine patch size to half the coarse one (or a quarter on GEOID-Flood) and the temporal resolutions to seasons and months, since sweeping these choices on every benchmark would be prohibitively expensive.
The best MiSTR pair of spatiotemporal resolutions is likely task-dependent and could benefit from hyperparameter tuning.

\textbf{Future work.}
Natural extensions include more than two levels of spatiotemporal resolution, and selectors that decide not only where and when to re-encode but also at which resolution, or even with which sensor.
Because the coarse tokens are cached, MiSTR could enable streaming computation for monitoring, where each new acquisition is encoded coarsely and re-encoded finely only where it changes.
This matters for recent Earth embedding products \citep{brown2025alphaearth,feng2026tessera,tseng2026olmoearthemb}, which are typically released at annual temporal resolution, partly because computing and storing embeddings at a finer fixed resolution is expensive.
MiSTR could re-encode only where and when a region changes, giving the users of such embeddings finer spatiotemporal resolution exactly where and when they need it.


\iftoggle{cvprfinal}{%
\section*{Acknowledgments}
GP and ES are supported by Canada CIFAR AI Chairs.
TZ is supported by a UBC 4YF fellowship and the Mitacs Accelerate program.
We acknowledge the support of the Natural Sciences and Engineering Research Council of Canada (NSERC: RGPIN-2024-06405, RGPIN-2025-06878).
Resources used in preparing this research were provided, in part, by the Province of Ontario, the Government of Canada through CIFAR, and companies sponsoring the Vector Institute.
}{}

{
    \small
    \bibliographystyle{ieeenat_fullname}
    \bibliography{references}

@string{iccv = "Proceedings of the International Conference on Computer Vision (ICCV)"}

@string{iclr = "International Conference on Learning Representations (ICLR)"}

@string{cvpr = "Proceedings of the IEEE Conference on Computer Vision and Pattern Recognition (CVPR)"}

@string{cvprw = "Proceedings of the IEEE Conference on Computer Vision and Pattern Recognition Workshops (CVPRW)"}

@string{wacv = "Proceedings of the IEEE/CVF Winter Conference on Applications of Computer Vision (WACV)"}

@string{neurips = "Advances in Neural Information Processing Systems (NeurIPS)"}

@string{neuripsdataset = "Advances in Neural Information Processing Systems (NeurIPS) Datasets and Benchmarks Track"}

@string{pami = "IEEE transactions on Pattern Analysis and Machine Intelligence (PAMI)"}

@string{icml = "International Conference on Machine Learning (ICML)"}

@string{eccvw = "Proceedings of the European Conference on Computer Vision Workshops (ECCVW)"}

@inproceedings{dosovitskiy2021an,
  title = {An Image is Worth 16x16 Words: Transformers for Image Recognition at Scale},
  author = {Alexey Dosovitskiy and Lucas Beyer and Alexander Kolesnikov and Dirk Weissenborn and Xiaohua Zhai and Thomas Unterthiner and Mostafa Dehghani and Matthias Minderer and Georg Heigold and Sylvain Gelly and Jakob Uszkoreit and Neil Houlsby},
  booktitle = iclr,
  year = {2021},
}

@article{szwarcman2025prithvi,
  title = {{Prithvi-EO-2.0}: A versatile multi-temporal foundation model for earth observation applications},
  author = {Szwarcman, Daniela and Roy, Sujit and Fraccaro, Paolo and G{\'\i}slason, Orsteinn El{\'\i} and Blumenstiel, Benedikt and Ghosal, Rinki and De Oliveira, Pedro Henrique and de Sousa Almeida, Joao Lucas and Sedona, Rocco and Kang, Yanghui and others},
  journal = {IEEE Transactions on Geoscience and Remote Sensing},
  year = {2026},
}

@inproceedings{cong2022satmae,
  title = {{SatMAE}: Pre-training transformers for temporal and multi-spectral satellite imagery},
  author = {Cong, Yezhen and Khanna, Samar and Meng, Chenlin and Liu, Patrick and Rozi, Erik and He, Yutong and Burke, Marshall and Lobell, David and Ermon, Stefano},
  booktitle = neurips,
  year = {2022},
}

@inproceedings{garnot2021panoptic,
  title = {Panoptic segmentation of satellite image time series with convolutional temporal attention networks},
  author = {Garnot, Vivien Sainte Fare and Landrieu, Loic},
  booktitle = iccv,
  year = {2021},
}

@inproceedings{beyer2023flexivit,
  title = {{FlexiViT}: One model for all patch sizes},
  author = {Beyer, Lucas and Izmailov, Pavel and Kolesnikov, Alexander and Caron, Mathilde and Kornblith, Simon and Zhai, Xiaohua and Minderer, Matthias and Tschannen, Michael and Alabdulmohsin, Ibrahim and Pavetic, Filip},
  booktitle = cvpr,
  year = {2023},
}

@inproceedings{nascetti2023biomassters,
  title = {{BioMassters}: A Benchmark Dataset for Forest Biomass Estimation using Multi-modal Satellite Time-series},
  author = {Andrea Nascetti and RITU YADAV and Kirill Brodt and Qixun Qu and Hongwei Fan and Yuri Shendryk and Isha Shah and Christine Chung},
  booktitle = neuripsdataset,
  year = {2023},
}

@inproceedings{miranda_2026_yieldsat,
  title = {{YieldSAT}: A Multimodal Benchmark Dataset for High-Resolution Crop Yield Prediction},
  author = {Miranda, Miro and Pathak, Deepak and Helber, Patrick and Bischke, Benjamin and Najjar, Hiba and Mena, Francisco and Sanchez, Cristhian and Pai, Akshay and Arenas, Diego and Toro, Matias Valdenegro and Charfuelan, Marcela and Nuske, Marlon and Dengel, Andreas},
  booktitle = cvpr,
  year = {2026},
}

@inproceedings{Chiriaco2026GEOIDFlood,
  title = {{GEOID-Flood}: A Large-Scale Multi-Modal Benchmark Dataset for Flood Segmentation},
  author = {Chiriaco, Gaetano and Barco, Luca and Bragagnolo, Andrea and Rossi, Claudio and Arnaudo, Edoardo},
  booktitle = eccvw,
  year = {2026},
}

@inproceedings{fu2024featup,
  title = {{FeatUp}: A model-agnostic framework for features at any resolution},
  author = {Fu, Stephanie and Hamilton, Mark and Brandt, Laura E and Feldmann, Axel and Zhang, Zhoutong and Freeman, William},
  booktitle = iclr,
  year = {2024},
}

@inproceedings{couairon2026jafar,
  title = {{JAFAR}: Jack up any feature at any resolution},
  author = {Couairon, Paul and Chambon, Loick and Serrano, Louis and Haugeard, Jean-Emmanuel and Cord, Matthieu and Thome, Nicolas},
  booktitle = neurips,
  year = {2025},
}

@inproceedings{wimmer2026anyup,
  title = {{AnyUp}: Universal Feature Upsampling},
  author = {Thomas Wimmer and Prune Truong and Marie-Julie Rakotosaona and Michael Oechsle and Federico Tombari and Bernt Schiele and Jan Eric Lenssen},
  booktitle = iclr,
  year = {2026},
}

@inproceedings{Seo_2026_CVPR,
  author = {Seo, Minseok and Hamilton, Mark and Kim, Changick},
  title = {Upsample Anything: A Simple and Hard to Beat Baseline for Feature Upsampling},
  booktitle = cvpr,
  year = {2026},
}

@inproceedings{ranzinger2025featsharp,
  title = {{FeatSharp}: Your vision model features, sharper},
  author = {Ranzinger, Mike and Heinrich, Greg and Molchanov, Pavlo and Catanzaro, Bryan and Tao, Andrew},
  booktitle = icml,
  year = {2025},
}

@article{planteur2026,
  title={{PLANTEUR}: A Crop-Type Segmentation Benchmark for Tropical Regions},
  author={Wolters, Piper and Tseng, Gabriel and Bastani, Favyen and Redmon, Joseph and Zhang, Yawen and Sablon, Hadrien and Shah, Ando},
  journal={arXiv preprint arXiv:2611.05223},
  year={2026}
}

@inproceedings{choudhury2025accelerating,
  title = {Accelerating vision transformers with adaptive patch sizes},
  author = {Choudhury, Rohan and Kim, JungEun and Park, Jinhyung and Yang, Eunho and Jeni, L{\'a}szl{\'o} A and Kitani, Kris M},
  booktitle = iclr,
  year = {2026},
}

@inproceedings{choi2024vid,
  title = {{vid-TLDR}: Training free token merging for light-weight video transformer},
  author = {Choi, Joonmyung and Lee, Sanghyeok and Chu, Jaewon and Choi, Minhyuk and Kim, Hyunwoo J},
  booktitle = cvpr,
  year = {2024},
}

@inproceedings{kirillov2020pointrend,
  title = {{PointRend}: Image segmentation as rendering},
  author = {Kirillov, Alexander and Wu, Yuxin and He, Kaiming and Girshick, Ross},
  booktitle = cvpr,
  year = {2020},
}

@inproceedings{rolf2024position,
  author = {Rolf, Esther and Klemmer, Konstantin and Robinson, Caleb and Kerner, Hannah},
  title = {Position: mission critical - satellite data is a distinct modality in machine learning},
  year = {2024},
  booktitle = icml,
}

@inproceedings{bolya2022token,
  title = {Token merging: Your {ViT} but faster},
  author = {Bolya, Daniel and Fu, Cheng-Yang and Dai, Xiaoliang and Zhang, Peizhao and Feichtenhofer, Christoph and Hoffman, Judy},
  booktitle = iclr,
  year = {2023},
}

@inproceedings{yun2019cutmix,
  title = {{CutMix}: Regularization Strategy to Train Strong Classifiers with Localizable Features},
  author = {Yun, Sangdoo and Han, Dongyoon and Oh, Seong Joon and Chun, Sanghyuk and Choe, Junsuk and Yoo, Youngjoon},
  booktitle = iccv,
  year = {2019},
}

@article{sykas2022sentinel,
  title = {A {Sentinel-2} multiyear, multicountry benchmark dataset for crop classification and segmentation with deep learning},
  author = {Sykas, Dimitrios and Sdraka, Maria and Zografakis, Dimitrios and Papoutsis, Ioannis},
  journal = {IEEE Journal of Selected Topics in Applied Earth Observations and Remote Sensing},
  year = {2022},
}

@inproceedings{Sani_2024_WACV,
  author = {Sani, Depanshu and Mahato, Sandeep and Saini, Sourabh and Agarwal, Harsh Kumar and Devshali, Charu Chandra and Anand, Saket and Arora, Gaurav and Jayaraman, Thiagarajan},
  title = {{SICKLE}: A Multi-Sensor Satellite Imagery Dataset Annotated With Multiple Key Cropping Parameters},
  booktitle = wacv,
  year = {2024},
}

@inproceedings{gerard2023wildfirespreadts,
  title = {WildfireSpread{TS}: A dataset of multi-modal time series for wildfire spread prediction},
  author = {Sebastian Gerard and Yu Zhao and Josephine Sullivan},
  booktitle = neuripsdataset,
  year = {2023},
}

@article{turkoglu2021crop,
  title = {Crop mapping from image time series: Deep learning with multi-scale label hierarchies},
  author = {Turkoglu, Mehmet Ozgur and D'Aronco, Stefano and Perich, Gregor and Liebisch, Frank and Streit, Constantin and Schindler, Konrad and Wegner, Jan Dirk},
  journal = {Remote Sensing of Environment},
  year = {2021},
}

@article{jakubik2023foundation,
  title = {Foundation models for generalist geospatial artificial intelligence},
  author = {Jakubik, Johannes and Roy, Sujit and Phillips, CE and Fraccaro, Paolo and Godwin, Denys and Zadrozny, Bianca and Szwarcman, Daniela and Gomes, Carlos and Nyirjesy, Gabby and Edwards, Blair and others},
  journal = {arXiv preprint arXiv:2310.18660},
  year = {2023},
}

@inproceedings{tseng2025galileo,
  title = {Galileo: Learning Global \& Local Features of Many Remote Sensing Modalities},
  author = {Tseng, Gabriel and Fuller, Anthony and Reil, Marlena and Herzog, Henry and Beukema, Patrick and Bastani, Favyen and Green, James R and Shelhamer, Evan and Kerner, Hannah and Rolnick, David},
  booktitle = icml,
  year = {2025},
}

@inproceedings{fuller2023croma,
  title = {{CROMA}: Remote sensing representations with contrastive radar-optical masked autoencoders},
  author = {Fuller, Anthony and Millard, Koreen and Green, James},
  booktitle = neurips,
  year = {2023},
}

@article{marsocci2024pangaea,
  title = {{PANGAEA}: Assessing Geospatial Foundation Models Capabilities through a Global and Inclusive Benchmark},
  author = {Marsocci, Valerio and Jia, Yuru and Bellier, Georges Le and Kerekes, David and Zeng, Liang and Hafner, Sebastian and Gerard, Sebastian and Brune, Eric and Yadav, Ritu and Shibli, Ali and others},
  journal = {IEEE Geoscience and Remote Sensing Magazine},
  year = {2026},
}

@inproceedings{liang2022not,
  title = {Not all patches are what you need: Expediting vision transformers via token reorganizations},
  author = {Liang, Youwei and Ge, Chongjian and Tong, Zhan and Song, Yibing and Wang, Jue and Xie, Pengtao},
  booktitle = iclr,
  year = {2022},
}

@inproceedings{herzog2026olmoearth,
  title = {{OlmoEarth}: Stable latent image modeling for multimodal earth observation},
  author = {Herzog, Henry and Bastani, Favyen and Zhang, Yawen and Tseng, Gabriel and Redmon, Joseph and Sablon, Hadrien and Park, Ryan and Morrison, Jacob and Buraczynski, Alexandra and Farley, Karen and others},
  booktitle = cvpr,
  year = {2026},
}

@article{salamatian2026lookwhen,
  title={LookWhen? Fast Video Recognition by Learning When, Where, and What to Compute},
  author={Salamatian, Ali and Fuller, Anthony and Sarkar, Pritam and Green, James R and Sigal, Leonid and Shelhamer, Evan},
  journal={arXiv preprint arXiv:2605.06809},
  year={2026}
}

@inproceedings{fuller2026lookwhere,
  title={LookWhere? {E}fficient Visual Recognition by Learning Where to Look and What to See from Self-Supervision},
  author={Fuller, Anthony and Yassin, Yousef and Wen, Junfeng and Ibrahim, Tarek and Kyrollos, Daniel and Green, James and Shelhamer, Evan},
  booktitle=neurips,
  year={2025}
}

@inproceedings{huang2022glance,
  title={Glance and focus networks for dynamic visual recognition},
  author={Huang, Gao and Wang, Yulin and Lv, Kangchen and Jiang, Haojun and Huang, Wenhui and Qi, Pengfei and Song, Shiji},
  booktitle=pami,
  year={2023},
}

@inproceedings{rao2021dynamicvit,
  title={{DynamicViT}: Efficient vision transformers with dynamic token sparsification},
  author={Rao, Yongming and Zhao, Wenliang and Liu, Benlin and Lu, Jiwen and Zhou, Jie and Hsieh, Cho-Jui},
  booktitle=neurips,
  year={2021}
}

@article{brown2025alphaearth,
  title={Alphaearth foundations: An embedding field model for accurate and efficient global mapping from sparse label data},
  author={Brown, Christopher F and Kazmierski, Michal R and Pasquarella, Valerie J and Rucklidge, William J and Samsikova, Masha and Zhang, Chenhui and Shelhamer, Evan and Lahera, Estefania and Wiles, Olivia and Ilyushchenko, Simon and others},
  journal={arXiv preprint arXiv:2507.22291},
  year={2025}
}

@inproceedings{feng2026tessera,
  title={{TESSERA}: Temporal embeddings of surface spectra for earth representation and analysis},
  author={Feng, Zhengpeng and Atzberger, Clement and Jaffer, Sadiq and Knezevic, Jovana and Sormunen, Silja and Young, Robin and Lisaius, Madeline C and Immitzer, Markus and Jackson, Toby and Ball, James and others},
  booktitle=cvpr,
  year={2026}
}

@article{tseng2026olmoearthemb,
  title={{OlmoEarth} v1.3: Embeddings},
  author={Tseng, Gabriel and Bastani, Favyen and Johnson, Patrick Alan and Zhang, Yawen and Wolters, Piper and Shah, Ando and Sablon, Hadrien and Redmon, Joseph and DeHaven, Heather and Ratner, Rachel and others},
  journal={arXiv preprint arXiv:2610.32294},
  year={2026}
}

@inproceedings{ronen2023vision,
  title={Vision transformers with mixed-resolution tokenization},
  author={Ronen, Tomer and Levy, Omer and Golbert, Avram},
  booktitle=cvprw,
  year={2023},
}

@inproceedings{jakubik2025terramind,
  title = {{TerraMind}: Large-Scale Generative Multimodality for Earth Observation},
  author = {Jakubik, Johannes and Yang, Felix and Blumenstiel, Benedikt and others},
  booktitle = iccv,
  year = {2025},
}

@inproceedings{dao2023flashattention2,
  title     = {{FlashAttention-2}: Faster Attention with Better Parallelism and Work Partitioning},
  author    = {Dao, Tri},
  booktitle = iclr,
  year      = {2024}
}

@inproceedings{han2021dynamic,
  title={Dynamic neural networks: A survey},
  author={Han, Yizeng and Huang, Gao and Song, Shiji and Yang, Le and Wang, Honghui and Wang, Yulin},
  booktitle=pami,
  year={2022},
}
}

\clearpage
\setcounter{page}{1}
\maketitlesupplementary
\appendix
\renewcommand{\dblfloatpagefraction}{0.8}
\noindent
\textbf{The supplementary material is organized as follows.}
\begin{itemize}
\item \textbf{Appendix~\ref{apx:separate}} shows how fast the token count grows with resolution, then varies spatial and temporal resolution separately: each helps SITS dense tasks on its own, but every step up costs several times more compute for a smaller gain.
\item \textbf{Appendix~\ref{apx:selection}} ablates the selectors: entropy (where) and decoder attention (when) beat random and grid selection.
\item \textbf{Appendix~\ref{apx:flexible-time-budget}} goes beyond selecting one season, and re-encodes up to two seasons per selected point, chosen at test time: at higher budgets, this lifts MiSTR above the finest fixed-resolution model.
\item \textbf{Appendix~\ref{apx:baseline-lr}} describes the baselines' learning-rate tuning; MiSTR and the fixed-resolution models use one untuned learning rate.
\item \textbf{Appendix~\ref{apx:throughput}} measures latency: MiSTR's GMAC savings carry over to wall-clock time, increasingly so with a larger backbone (MiSTR OlmoEarth Base).
\item \textbf{Appendix~\ref{apx:selection-samples}} visualizes spatiotemporal selection: MiSTR re-encodes uncertain boundaries in informative seasons, and spends more tokens on complex scenes.
\item \textbf{Appendix~\ref{apx:one-way-attention-ablation}} compares one-way attention with self- and cross-attention: one-way attention gives up almost no performance of full self-attention at just 57--71\% of its compute; its \hrtok{} tokens use both \lrtok{} and \hrtok{} context, and the decoder combines both passes into a prediction better than either.
\item \textbf{Appendix~\ref{apx:alt-rsfm-init}} initializes MiSTR from TerraMind (space-only) and Tessera v2 (time-only): MiSTR is compatible with existing RSFMs other than OlmoEarth.
\item \textbf{Appendix~\ref{apx:matched-compute}} compares MiSTR against the Pareto front of fixed-resolution models at matched accuracy and matched compute: MiSTR needs 1.3--4.6$\times$ fewer GMACs, and is more accurate than 28 of the 34 Pareto-optimal models at no more compute.
\item \textbf{Appendix~\ref{apx:seeds}} presents 3-seed standard errors from our main results on the many-input one-output benchmarks of Fig.~\ref{fig:result}, and shows they are small compared with MiSTR's gains.
\item \textbf{Appendix~\ref{apx:geoid-flood}} details flood-tracking training (with temporal CutMix) and evaluation: MiSTR's compute saving grows with history length. We further present representative sample flood series to show qualitatively why MiSTR is well-fitted for this task. Visualizations show MiSTR can track flood events of various length.
\item \textbf{Appendix~\ref{apx:datasets}} summarizes each dataset's task, inputs, targets and splits, with its series length and one sample.
\end{itemize}

\section{Separating Spatial \& Temporal Resolution}
\label{apx:separate}
MiSTR adds resolution along two axes, smaller patches (space) and more timesteps (time), but only where and when its selectors ask for it.
Here we verify that finer resolution along each axis improves performance by training fixed-resolution OlmoEarth with spatial and temporal resolution varied separately.
Each axis improves accuracy on its own, and each step up costs several times more compute for a smaller gain.
A fixed-resolution model pays this rising cost at every location and timestep of a tile; MiSTR pays it only at the points and seasons it selects.

\begin{table}[t]
\centering
\caption{\textbf{Tokens per SITS} for fixed-resolution OlmoEarth on the $64\times64$-pixel tiles of the four main datasets. Fewer tokens on YieldSAT: Sentinel-2 only, and fewer timesteps with observations. $S_2T_\text{sea}$ and $S_4T_\text{mon}$ are scaled from the $S_{16}$ counts by the number of patches. For reference, a typical ViT image classifier processes an ImageNet sample as 196 or 256 tokens, up to two orders of magnitude fewer.}
\label{tab:token-counts}
\footnotesize
\setlength{\tabcolsep}{4pt}
\begin{tabular}{lrrrr}
\toprule
 & PASTIS & PLANTEUR & BioMassters & YieldSAT \\
\midrule
$S_{16}T_\text{sea}$ & 256 & 256 & 256 & 142 \\
$S_{12}T_\text{sea}$ & 576 & 576 & 576 & 320 \\
$S_{8}T_\text{sea}$ & 1{,}024 & 1{,}024 & 1{,}024 & 568 \\
$S_{6}T_\text{sea}$ & 1{,}936 & 1{,}936 & 1{,}936 & 1{,}075 \\
$S_{4}T_\text{sea}$ & 4{,}096 & 4{,}096 & 4{,}096 & 2{,}274 \\
$S_{2}T_\text{sea}$ & 16{,}384 & 16{,}384 & 16{,}384 & 9{,}096 \\
\midrule
$S_{16}T_\text{mon}$ & 720 & 768 & 672 & 321 \\
$S_{12}T_\text{mon}$ & 1{,}620 & 1{,}728 & 1{,}512 & 723 \\
$S_{8}T_\text{mon}$ & 2{,}880 & 3{,}072 & 2{,}688 & 1{,}286 \\
$S_{6}T_\text{mon}$ & 5{,}445 & 5{,}808 & 5{,}082 & 2{,}430 \\
$S_{4}T_\text{mon}$ & 11{,}520 & 12{,}288 & 10{,}752 & 5{,}144 \\
\bottomrule
\end{tabular}
\end{table}

\subsection{Spatial resolution}
\begin{figure}[t]
    \centering
    \includegraphics[width=\linewidth]{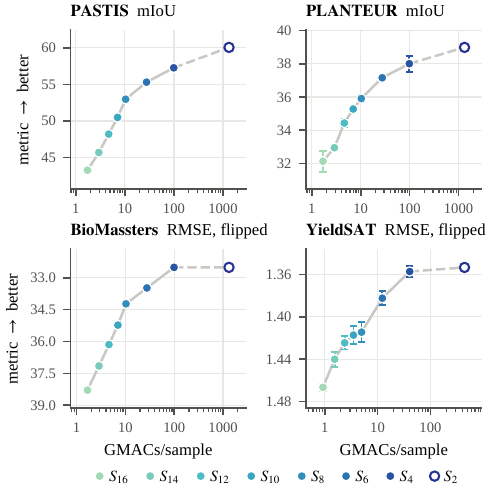}
    \caption{\textbf{Higher spatial resolution helps, with diminishing returns.}
    Fixed-resolution OlmoEarth-Tiny on seasonal composites ($T_\text{sea}$) with patch sizes from 16 to 2 pixels.
    Means over three seeds with $\pm1$ standard error; where no bar is visible, it is smaller than the marker.
    Due to its high computational cost, $S_2$ (hollow) is a single run trained for half the epochs.
    RMSE axes are flipped so that up is better.}
    \label{fig:patch_size_ablation}
\end{figure}
\Cref{fig:patch_size_ablation} varies the patch size from 16 to 2 pixels with seasonal composites as input.
Smaller patches improve accuracy on every dataset, but the gains shrink as the cost grows: on PASTIS, going from $S_{16}$ to $S_8$ gains $9.7{\pm}0.2$ mIoU for $6\times$ the GMACs, and $S_8$ to $S_4$ gains $4.3{\pm}0.3$ for a further $9.6\times$.
PLANTEUR follows the same pattern ($+3.8{\pm}0.7$, then $+2.1{\pm}0.5$ mIoU), as does BioMassters, whose RMSE drops by $4.06{\pm}0.02$, then $1.72{\pm}0.05$.
On YieldSAT, both steps lower RMSE by similar amounts ($0.052{\pm}0.010$ and $0.057{\pm}0.011$).
$S_2$ costs another $11$--$13\times$ the GMACs of $S_4$ but adds at most 2.8 mIoU (PASTIS) and almost nothing on BioMassters and YieldSAT.

\subsection{Temporal resolution}
\begin{figure}[t]
    \centering
    \includegraphics[width=\linewidth]{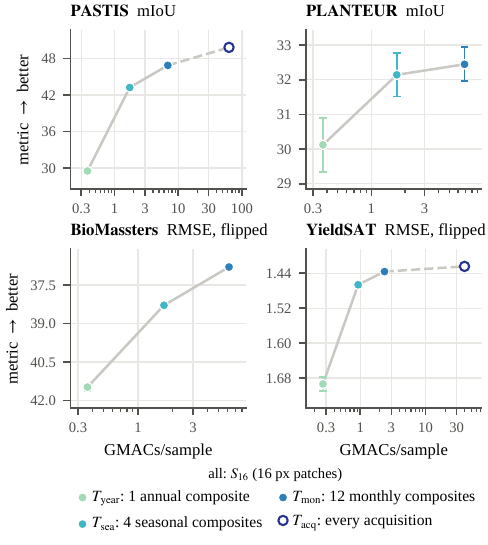}
    \caption{\textbf{Higher temporal resolution helps, with diminishing returns.}
    Fixed-resolution OlmoEarth-Tiny with 16-pixel patches ($S_{16}$) on one annual ($T_\text{year}$), four seasonal ($T_\text{sea}$) or twelve monthly ($T_\text{mon}$) composites, or on every acquisition ($T_\text{acq}$).
    All composite inputs are trained as in \cref{fig:result}.
    Means over three seeds with $\pm1$ standard error; where no bar is visible, it is smaller than the marker.
    Due to its high computational cost, $T_\text{acq}$ (hollow; PASTIS and YieldSAT) is a single run trained for half the epochs.
    RMSE axes are flipped so that up is better.}
    \label{fig:temporal_resolution_ablation}
\end{figure}
\Cref{fig:temporal_resolution_ablation} holds the patch size at 16 pixels and refines the input from one annual composite to every acquisition.
Each finer cadence improves accuracy, again with diminishing returns.
Going from one composite to four seasonal composites gives the largest gain on every dataset ($13.7{\pm}0.2$ mIoU on PASTIS, $2.0{\pm}1.0$ mIoU on PLANTEUR, $3.19{\pm}0.14$ RMSE on BioMassters, $0.227{\pm}0.015$ RMSE on YieldSAT), and monthly composites add less ($3.6{\pm}0.2$ mIoU, $0.3{\pm}0.8$ mIoU, $1.49{\pm}0.03$ RMSE, $0.030{\pm}0.004$ RMSE) for $2.5$--$4\times$ the GMACs.
Using every acquisition adds 3.0 mIoU on PASTIS and 0.012 RMSE on YieldSAT, for $9$--$17\times$ the GMACs of monthly composites.
MiSTR therefore encodes its context pass from seasonal composites and adds monthly composites in its detail pass only where they are needed (\cref{sec:results}).

\subsection{Token counts}
\label{apx:token-counts}
SITS are long token sequences, which is what makes fine resolution everywhere expensive (\cref{sec:intro}).
A remote sensing instance is tokenized along space, time and modality: OlmoEarth turns every band group of every modality, at every timestep and patch location, into one token (\cref{sec:existing_work}).
The token count is therefore the product of the number of patches, timesteps and band groups; PASTIS, PLANTEUR and BioMassters have four band groups (three of Sentinel-2, one of Sentinel-1), YieldSAT three.
\Cref{tab:token-counts} lists the resulting counts.
Even on $64\times64$-pixel tiles, a monthly series at 4-pixel patches has over 10{,}000 tokens on three of the four datasets, and a seasonal series at 2-pixel patches over 16{,}000; larger tiles multiply these counts by their area.
On GEOID-Flood, a 32-timestep series of $512\times512$-pixel Sentinel-1 tiles has 8{,}192 tokens at a 32-pixel patch size.

\FloatBarrier

\section{Separating Spatial \& Temporal Selection}
\label{apx:selection}
\begin{table*}[t]
\centering
\caption{\textbf{Where to re-encode.} PASTIS mIoU and compute at equal \hrtok{} budget: $n$ points per sample (on average for entropy, which uses one global threshold), i.e.\ a share $q$ of all \Shi{} points, each re-encoded at the months of the season picked by decoder attention. Only \emph{where} differs. Mean $\pm$ standard error over three seeds; GMACs vary across seeds because selection is adaptive (\cref{apx:selection}).}
\label{tab:selector}
\setlength{\tabcolsep}{6pt}
\begin{tabular}{lcccccc}
\toprule
 & \multicolumn{3}{c}{$S_{16}T_\text{sea}$ \& $S_8T_\text{mon}$} & \multicolumn{3}{c}{$S_8T_\text{sea}$ \& $S_4T_\text{mon}$} \\
\cmidrule(lr){2-4} \cmidrule(lr){5-7}
$n$ & 8 & 16 & 32 & 32 & 64 & 128 \\
$q$ & 1/8 & 1/4 & 1/2 & 1/8 & 1/4 & 1/2 \\
\midrule
\multicolumn{7}{l}{\emph{test mIoU}} \\
no \hrtok{} & 43.3\,{\footnotesize$\pm$0.2} & 43.3\,{\footnotesize$\pm$0.2} & 43.3\,{\footnotesize$\pm$0.2} & 53.0\,{\footnotesize$\pm$0.1} & 53.0\,{\footnotesize$\pm$0.1} & 53.0\,{\footnotesize$\pm$0.1} \\
random & 44.7\,{\footnotesize$\pm$0.2} & 45.3\,{\footnotesize$\pm$0.1} & 47.5\,{\footnotesize$\pm$0.2} & 53.4\,{\footnotesize$\pm$0.2} & 53.7\,{\footnotesize$\pm$0.2} & 54.6\,{\footnotesize$\pm$0.1} \\
grid & 44.6\,{\footnotesize$\pm$0.1} & 45.5\,{\footnotesize$\pm$0.2} & 47.5\,{\footnotesize$\pm$0.1} & 53.6\,{\footnotesize$\pm$0.2} & 53.9\,{\footnotesize$\pm$0.2} & 54.7\,{\footnotesize$\pm$0.2} \\
entropy & \textbf{47.4}\,{\footnotesize$\pm$0.2} & \textbf{49.7}\,{\footnotesize$\pm$0.0} & \textbf{51.6}\,{\footnotesize$\pm$0.2} & \textbf{55.4}\,{\footnotesize$\pm$0.5} & \textbf{56.1}\,{\footnotesize$\pm$0.1} & \textbf{57.2}\,{\footnotesize$\pm$0.2} \\
\midrule
\multicolumn{7}{l}{\emph{GMACs per SITS (test)}} \\
no \hrtok{} & 1.73\,{\footnotesize$\pm$0.00} & 1.73\,{\footnotesize$\pm$0.00} & 1.73\,{\footnotesize$\pm$0.00} & 10.37\,{\footnotesize$\pm$0.00} & 10.37\,{\footnotesize$\pm$0.00} & 10.37\,{\footnotesize$\pm$0.00} \\
random & 2.48\,{\footnotesize$\pm$0.00} & 3.26\,{\footnotesize$\pm$0.00} & 5.07\,{\footnotesize$\pm$0.00} & 14.97\,{\footnotesize$\pm$0.02} & 20.82\,{\footnotesize$\pm$0.06} & 36.72\,{\footnotesize$\pm$0.10} \\
grid & 2.48\,{\footnotesize$\pm$0.00} & 3.26\,{\footnotesize$\pm$0.00} & 5.07\,{\footnotesize$\pm$0.00} & 14.97\,{\footnotesize$\pm$0.02} & 20.82\,{\footnotesize$\pm$0.06} & 36.75\,{\footnotesize$\pm$0.08} \\
entropy & 2.63\,{\footnotesize$\pm$0.06} & 3.45\,{\footnotesize$\pm$0.07} & 5.27\,{\footnotesize$\pm$0.08} & 16.05\,{\footnotesize$\pm$0.26} & 22.20\,{\footnotesize$\pm$0.12} & 38.46\,{\footnotesize$\pm$0.38} \\
\bottomrule
\end{tabular}
\end{table*}

\begin{table*}[t]
\centering
\caption{\textbf{When to re-encode.} PASTIS mIoU and compute at equal \hrtok{} budget: the same entropy-picked points ($q = 1/4$), each re-encoded at the months of one season; only \emph{which} season differs (grid: seasons cycle over the lattice of points). Mean $\pm$ standard error over three seeds; GMACs vary across seeds because selection is adaptive (\cref{apx:selection}).}
\label{tab:time-selector}
\setlength{\tabcolsep}{6pt}
\begin{tabular}{lcccc}
\toprule
 & \multicolumn{2}{c}{$S_{16}T_\text{sea}$ \& $S_8T_\text{mon}$} & \multicolumn{2}{c}{$S_8T_\text{sea}$ \& $S_4T_\text{mon}$} \\
\cmidrule(lr){2-3} \cmidrule(lr){4-5}
 & test mIoU & GMACs & test mIoU & GMACs \\
\midrule
no \hrtok{} & 43.3\,{\footnotesize$\pm$0.2} & 1.73\,{\footnotesize$\pm$0.00} & 53.0\,{\footnotesize$\pm$0.1} & 10.37\,{\footnotesize$\pm$0.00} \\
\midrule
random & 47.2\,{\footnotesize$\pm$0.2} & 3.34\,{\footnotesize$\pm$0.01} & 55.8\,{\footnotesize$\pm$0.1} & 21.89\,{\footnotesize$\pm$0.04} \\
grid & 47.6\,{\footnotesize$\pm$0.1} & 3.34\,{\footnotesize$\pm$0.01} & 55.9\,{\footnotesize$\pm$0.2} & 21.82\,{\footnotesize$\pm$0.09} \\
per-season entropy & 47.4\,{\footnotesize$\pm$0.2} & 3.42\,{\footnotesize$\pm$0.01} & 55.3\,{\footnotesize$\pm$0.3} & 22.01\,{\footnotesize$\pm$0.13} \\
attention & \textbf{49.7}\,{\footnotesize$\pm$0.0} & 3.45\,{\footnotesize$\pm$0.07} & \textbf{56.1}\,{\footnotesize$\pm$0.1} & 22.20\,{\footnotesize$\pm$0.12} \\
\bottomrule
\end{tabular}
\end{table*}

MiSTR selects with two simple heuristic signals (\cref{subsec:selection}); here we show that each picks better than an image-agnostic choice.
Selection splits into two decisions: \emph{where} to add the fine patch size, chosen by the entropy or the variance of the context pass's prediction averaged over the pixels of each candidate point, and \emph{when}, chosen by the decoder's attention over seasons.
We test each signal on PASTIS with the other axis held fixed, against two baselines that ignore the image: random and regular-grid selection of \hrtok{} points.
Re-encoding the selected points improves on both baselines in every configuration: entropy-picked points by 1.7--4.2 mIoU, attention-picked seasons by 0.2--2.1 mIoU.
Every model follows the protocol of \cref{sec:results}: 50 epochs of guided warm-up, then 50 epochs with its own selector.
All numbers are mean $\pm$ standard error over three seeds, compute included: selection is adaptive, so each trained checkpoint re-encodes a different number of tokens on the same test set, and its GMACs differ from seed to seed (the fixed-resolution model's do not).

\paragraph{Where.}
\Cref{tab:selector} holds time fixed at MiSTR's choice: each selected point re-encodes the season picked by decoder attention, and methods differ only in which $n$ points they pick.
Random draws the $n$ points uniformly over the tile; grid places them on a regular lattice spread evenly over the tile (with a random offset during training and a fixed one at test).
Entropy beats both baselines at every budget for both resolution pairs: by $2.7{\pm}0.2$ to $4.2{\pm}0.2$ mIoU over the better of the two with $S_{16}T_\text{sea}$ \& $S_8T_\text{mon}$, and by $1.7{\pm}0.6$ to $2.5{\pm}0.2$ with $S_8T_\text{sea}$ \& $S_4T_\text{mon}$.
The margin is larger at larger budgets.
Random and grid recover only 20--34\% of entropy's gain over no \hrtok{} tokens at one eighth of the points, and 39--51\% at one half: an image-agnostic pick spends most of its budget where the context pass is already right.
Entropy at $n$ matches both baselines at $4n$ ($-0.2{\pm}0.3$ mIoU with $S_{16}T_\text{sea}$ \& $S_8T_\text{mon}$, $+0.7{\pm}0.5$ with $S_8T_\text{sea}$ \& $S_4T_\text{mon}$) at 52\% and 44\% of their GMACs.

\paragraph{When.}
\Cref{tab:time-selector} holds space fixed at MiSTR's choice: the same entropy-picked points ($q{=}1/4$) each re-encode one season, and methods differ only in which.
Random picks a uniformly random season per point; grid cycles the seasons over the lattice of points.
Decoder attention is best at both resolution pairs.
It leads the best baseline by $2.1{\pm}0.1$ mIoU with $S_{16}T_\text{sea}$ \& $S_8T_\text{mon}$, and by a smaller $0.2{\pm}0.2$ (grid) and $0.3{\pm}0.2$ (random) with $S_8T_\text{sea}$ \& $S_4T_\text{mon}$, whose finer context pass already resolves most of what a single extra season adds.
Per-season entropy, the spatial signal applied to time, is no better than random.
The decoder is trained only on its mixture over seasons, so each season's own prediction, and hence its entropy, is not trained to be accurate on its own, whereas the attention weights are exactly how much the prediction draws on each season.
\Cref{apx:flexible-time-budget} studies re-encoding more than one season per selected point.

\section{Flexible Temporal Selection Budget}
\label{apx:flexible-time-budget}
\begin{figure}[t]
    \centering
    \includegraphics[width=\linewidth]{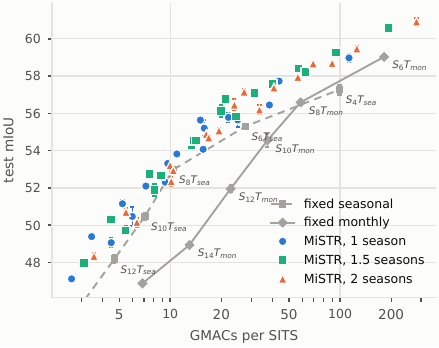}
    \caption{\textbf{Flexible temporal selection budget} on PASTIS (test mIoU vs.\ GMACs per SITS). Each MiSTR checkpoint (six resolution pairs $\times$ $q \in \{1/8, 1/4, 1/2\}$) is trained with 1, 1.5 or 2 seasons per selected point, drawn at random per batch, and tested with each; 1.5 seasons gives a second season to the most uncertain half of the selected points. Gray: fixed-resolution OlmoEarth models. Error bars: $\pm 1$ standard error over three seeds.}
    \label{fig:flexible-time-budget}
\end{figure}
MiSTR re-encodes each selected point from a single season, the one its decoder attends to most (\cref{subsec:selection}).
How many seasons to re-encode per point is, however, a choice: a second season is another way to spend compute, alongside selecting more points.
We therefore train MiSTR so that this choice can be left to the practitioner at inference time.
In each training batch, a checkpoint re-encodes 1, 1.5 or 2 seasons per selected point, drawn at random; at 1.5 seasons, only the most uncertain half of the selected points gets a second season.
The second season is the one whose own \lrtok{} prediction disagrees most with that of the most attended season (Jensen--Shannon divergence).
At test time, the same checkpoint runs with 1, 1.5 or 2 seasons per point, with the threshold refit on the validation set.
We train six PASTIS pairs, $S_pT_\text{sea}$ \& $S_{p/2}T_\text{mon}$ for $p \in \{16, 14, 12, 10, 8, 6\}$, at $q \in \{1/8, 1/4, 1/2\}$ with the recipe of \cref{sec:results}.

Different configurations give the best accuracy at different computational budgets (\cref{fig:flexible-time-budget}).
At low budgets, one season per point is competitive.
Once more compute is available than one season can use, a second season on the most uncertain half of the points (1.5 seasons) spends it most effectively: it lifts MiSTR above the finest fixed-resolution model, with $S_6T_\text{sea}$ \& $S_3T_\text{mon}$ ($q = 1/2$) reaching 60.6 mIoU at 193 GMACs against 59.0 for $S_6T_\text{mon}$ at 182 GMACs.
Whichever configuration is chosen, MiSTR beats fixed resolution: 52 of the 54 MiSTR points lie above both the seasonal and the monthly fixed-resolution curves.
\FloatBarrier

\section{Baseline Learning Rates}
\label{apx:baseline-lr}
The baselines of \cref{fig:baselines} come from image and video processing and have no established learning rate for SITS, so we tune each trained baseline on the validation set.
For APT and AnyUp, we sweep $\{3\cdot10^{-4}, 10^{-3}, 3\cdot10^{-3}, 10^{-2}\}$ on one representative configuration ($S_{12}T_\text{sea}$) on PASTIS; both pick $3\cdot10^{-3}$, which we reuse for the remaining configurations and for YieldSAT.
The bilinear head that reads Upsample Anything's upsampled features is swept over $\{10^{-4}, 10^{-3}, 10^{-2}\}$ for every patch size on PASTIS and picks $10^{-2}$ at $S_{16}$ and $S_{12}$ and $10^{-3}$ at $S_8$; on YieldSAT it is trained at $10^{-2}$.
vid-TLDR is training-free.
MiSTR and the fixed-resolution models use a learning rate of $10^{-3}$ throughout, without tuning.

\section{Measuring MiSTR Throughput}
\label{apx:throughput}
\begin{figure*}[t]
    \centering
    \includegraphics[width=\linewidth]{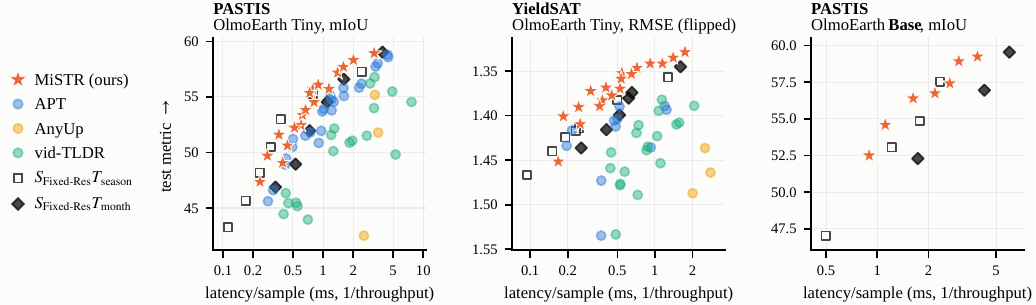}
    \caption{\textbf{MiSTR is fast in wall-clock time, and even more so as the backbone grows.} Test metric vs.\ measured latency per sample (1/throughput) on one NVIDIA L40S, each model at its fastest batch size of 32, 64, 128 or 256. Left and middle: OlmoEarth Tiny on PASTIS and YieldSAT (RMSE axis flipped, so up is better), with the baselines of \cref{fig:baselines} (Upsample Anything, a test-time optimization method, is not shown: it is much slower than all other models and less accurate on these tasks). Right: MiSTR and fixed-resolution models trained from OlmoEarth Base on PASTIS ($S_{16}T_\text{sea}$ \& $S_8T_\text{mon}$, $S_{12}T_\text{sea}$ \& $S_6T_\text{mon}$, $S_{10}T_\text{sea}$ \& $S_5T_\text{mon}$). Fixed-resolution models slower than twice the slowest MiSTR model are not shown.}
    \label{fig:throughput}
\end{figure*}
In \cref{subsec:baselines} we showed that MiSTR gives a favorable accuracy--compute tradeoff compared to fixed-resolution models and to baselines from image and video processing, measured in GMACs.
GMACs, however, do not always translate into wall-clock time: irregular operations such as gathering selected tokens, sorting, or attention without fused kernels can make a method slower than its GMACs suggest.
Here we show that MiSTR keeps its lead over the baselines of \cref{fig:baselines} in measured latency, and that its lead over fixed-resolution OlmoEarth grows with the backbone.

\paragraph{Measurement.}
We measure the inference throughput of every trained model on the test split on a single NVIDIA L40S GPU (48\,GB), in bfloat16 mixed precision, with the input series already on the GPU so that data loading is excluded.
Because the fastest batch size differs between models, we sweep batch sizes of 32, 64, 128 and 256 for each model and report its highest throughput, over 1024 test samples and the best of three repetitions after a warm-up pass.
For MiSTR, the timed pass includes everything at inference: the \lrtok{} pass, selection against the validation-set threshold, tokenizing the selected points, the one-way \hrtok{} pass, and the decoder.
\Cref{fig:throughput} reports latency per sample, the inverse of throughput.

\paragraph{Why MiSTR is fast.}
Two design choices keep MiSTR's extra operations cheap on a GPU.
First, selection compares each point's uncertainty to a single precomputed threshold (\cref{subsec:selection}), an elementwise operation; it needs no per-sample ranking or top-$k$ of the uncertainty, and no learned scoring network.
Second, one-way attention (\cref{subsec:one-way-attention}) is a standard attention in which the \hrtok{} tokens are the only queries and the cached \lrtok{} keys and values of the same layer are concatenated with those of the \hrtok{} tokens.
It therefore needs no attention mask and runs with FlashAttention \citep{dao2023flashattention2}, packing samples with different numbers of selected points into one variable-length call, and the \lrtok{} keys and values are reused from the first pass rather than recomputed.
By contrast, vid-TLDR scores tokens with explicit attention maps, which rules out fused attention kernels; on PASTIS it needs 3.3\,ms per sample to reach 56.8 mIoU, while MiSTR reaches 57.3 mIoU at 1.4\,ms.

\paragraph{OlmoEarth Tiny.}
At coarse resolutions, OlmoEarth Tiny is so small that its inference is limited by the CPU launching GPU kernels rather than by GPU compute, which penalizes MiSTR's two passes relative to a single fixed-resolution pass.
On PASTIS, MiSTR therefore lies on the seasonal fixed-resolution curve in latency at low resolution; as resolution increases, it lies above the monthly fixed-resolution curve and ahead of the baselines.
Even in this regime, MiSTR outperforms fixed-resolution models on YieldSAT at every latency: $S_{16}T_\text{sea}$ \& $S_8T_\text{mon}$ reaches an RMSE of 1.389 at 0.18\,ms per sample, against 1.409 for fixed $S_{10}T_\text{sea}$ at 0.23\,ms, and $S_6T_\text{sea}$ \& $S_3T_\text{mon}$ reaches 1.321 at 1.40\,ms, against 1.337 for fixed $S_6T_\text{mon}$ at 1.60\,ms.

\paragraph{OlmoEarth Base.}
We therefore also train MiSTR from OlmoEarth Base on PASTIS, with the recipe of \cref{sec:results}, to test a backbone large enough to be GPU-bound.
With Base, MiSTR's wall-clock advantage over fixed-resolution OlmoEarth in the low-spatiotemporal resolution regime becomes clearer (\cref{fig:throughput}, right).
$S_{16}T_\text{sea}$ \& $S_8T_\text{mon}$ ($q = 1/4$) reaches 54.6 mIoU at 1.12\,ms per sample, more accurate and faster than fixed $S_{12}T_\text{sea}$ (53.0 at 1.21\,ms), and $S_{10}T_\text{sea}$ \& $S_5T_\text{mon}$ ($q = 1/8$) reaches 58.9 mIoU at 3.02\,ms, above fixed $S_{12}T_\text{mon}$ (56.9 at 4.27\,ms) and within 0.6 mIoU of fixed $S_{10}T_\text{mon}$ at half its latency (59.5 at 5.98\,ms).
Once GPU time dominates, MiSTR's GMAC savings translate into wall-clock savings over fixed-resolution OlmoEarth.
\FloatBarrier

\section{MiSTR Spatiotemporal Selection Samples}
\label{apx:selection-samples}
\begin{figure*}[p]
    \centering
    \includegraphics[width=0.94\linewidth]{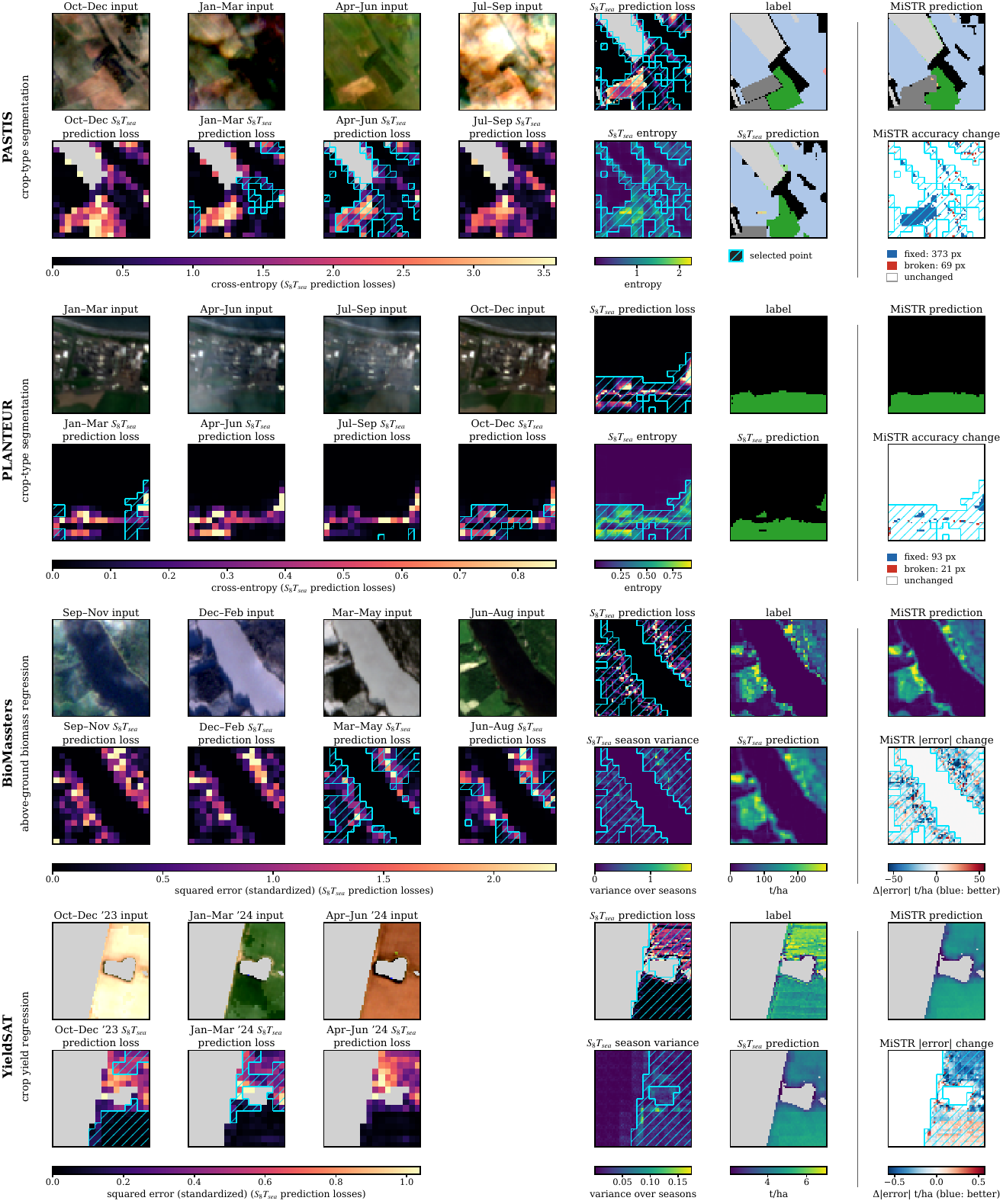}
    \caption{\textbf{MiSTR selection} ($S_8T_\text{sea}$ \& $S_4T_\text{mon}$, $q = 1/2$) on one test tile per dataset (rows).
    Top row of each dataset: each season's Sentinel-2 composite (RGB), the $S_8T_\text{sea}$ (\lrtok{}) prediction loss, the label, and MiSTR's prediction.
    Bottom row: the loss of each season token's own prediction, the $S_8T_\text{sea}$ uncertainty, the $S_8T_\text{sea}$ prediction, and where MiSTR's prediction became right or wrong.
    Hatched: selected points; in a season's loss panel, the points that selected that season. Gray: unlabeled.
    Regression (BioMassters, YieldSAT): squared error as the loss, the variance of the season tokens' predictions as the uncertainty, and the change in absolute error (blue: MiSTR closer to the label); composites are contrast-stretched per season. YieldSAT shows only the three seasons with observations.}
    \label{fig:selection-samples}
\end{figure*}

To show what the two selectors of \cref{subsec:selection} pick in practice, \cref{fig:selection-samples} shows where and when MiSTR $S_8T_\text{sea}$ \& $S_4T_\text{mon}$ ($q = 1/2$) re-encodes \hrtok{} tokens on one test tile of each dataset.
The spatial selection follows the uncertainty of the \lrtok{} prediction (entropy for classification, the variance of the season tokens' predictions for regression), which concentrates on field boundaries and on the regions the \lrtok{} prediction gets wrong.
For each selected point, the temporal selection mostly picks a season whose own prediction loss there is low, \ie{} the season most informative for that location, and MiSTR's second pass fixes far more pixels than it breaks (373 vs.\ 69 on PASTIS, 93 vs.\ 21 on PLANTEUR).

The selected points cover the uncertain parcel boundaries on PASTIS, the uncertain boundary of the labeled crop field on PLANTEUR, the forest edges along the river on BioMassters, and the field edges on YieldSAT.
Their monthly tokens come from Jan--Mar and Apr--Jun on PASTIS, mostly from Oct--Dec and Jan--Mar on PLANTEUR and YieldSAT, and from Mar--May and Jun--Aug on BioMassters, avoiding the snow-covered Dec--Feb.

\begin{figure*}[t]
    \centering
    \includegraphics[width=\linewidth]{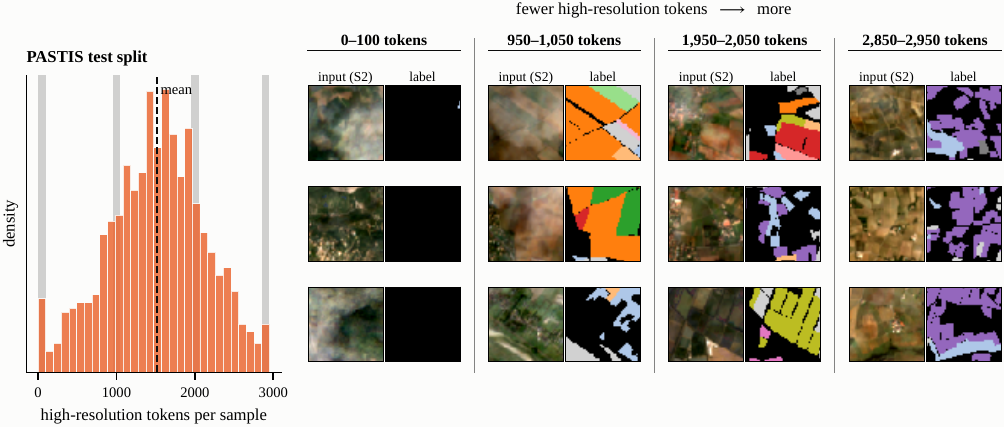}\\[6pt]
    \includegraphics[width=\linewidth]{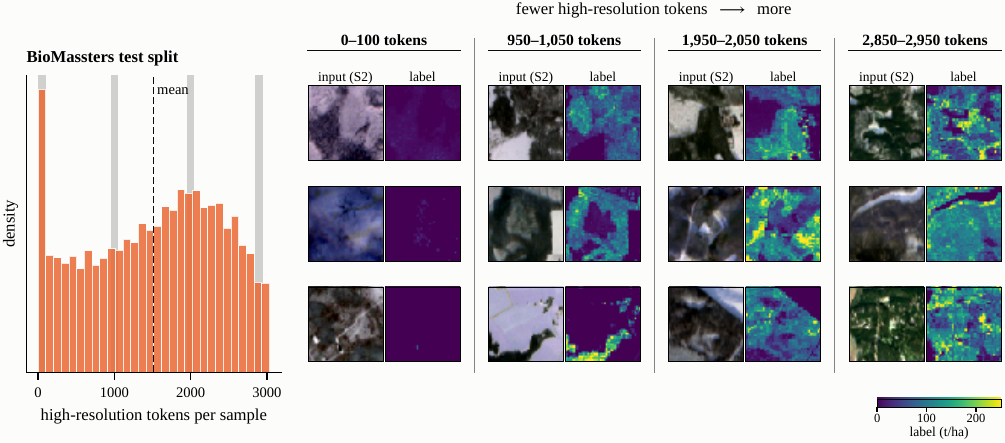}
    \caption{\textbf{MiSTR spends more \hrtok{} tokens on more complex samples.} MiSTR $S_8T_\text{sea}$ \& $S_4T_\text{mon}$ ($q = 1/2$, OlmoEarth Tiny) on the PASTIS (top) and BioMassters (bottom) test splits. Left: distribution of the number of \hrtok{} tokens per sample (dashed: mean). Right: three test samples drawn at random from each gray bin of that distribution (the token ranges above the columns), from few to many tokens, each shown as its Sentinel-2 input (RGB, per-pixel median of the seasonal composites) next to its label. PASTIS labels in the class colors of \cref{fig:selection-samples}, BioMassters labels in t/ha; gray: unlabeled.}
    \label{fig:per-sample-compute}
\end{figure*}
\subsection{Per-sample compute}
\label{apx:per-sample-compute}
For MiSTR $S_8T_\text{sea}$ \& $S_4T_\text{mon}$ ($q = 1/2$, OlmoEarth Tiny), the model of \cref{fig:selection-samples}, we measure how many \hrtok{} tokens each test sample receives.
MiSTR's budget $q$ sets the share of selected points over the whole validation set, not per sample: a point is selected when its uncertainty exceeds a single threshold (\cref{subsec:selection}).
The number of \hrtok{} tokens therefore varies from sample to sample, and \cref{fig:per-sample-compute} shows how it is distributed and which samples receive few or many.

On PASTIS, a test sample receives $1{,}516 \pm 6$ \hrtok{} tokens on average, with the 5th and 95th percentiles at $398 \pm 14$ and $2{,}515 \pm 34$ tokens; on BioMassters, the average is $1{,}518 \pm 3$ and the 5th and 95th percentiles are $44 \pm 4$ and $2{,}809 \pm 1$ (mean $\pm$ standard error over three seeds; the figure shows seed 0).
The samples that receive the fewest tokens are simple scenes: PASTIS tiles covered by background alone and BioMassters tiles of uniformly low biomass.
The samples that receive the most are complex: PASTIS tiles of many small parcels of different crops and BioMassters tiles where dense forest alternates with clearings.
MiSTR thus spends its compute where the \lrtok{} pass is uncertain, which is where the scene is complex.
The latency measurements of \cref{apx:throughput} run on batches of test samples, so they already include this per-sample variation.
\FloatBarrier

\section{One-Way Attention: Ablation and Analysis}
\label{apx:one-way-attention-ablation}
MiSTR's second pass uses one-way attention (\cref{subsec:one-way-attention}); here we compare it with the two standard alternatives and analyze what the trained model does with it.
Self-attention re-encodes every \lrtok{} token jointly with the selected \hrtok{} tokens.
One-way attention reuses the cached \lrtok{} pass and runs only the \hrtok{} tokens through the blocks, letting them read the cached \lrtok{} tokens and each other.
Cross-attention is cheaper still, but its \hrtok{} tokens read only the cached \lrtok{} tokens, not each other.
One-way attention keeps the accuracy of self-attention at a large saving in computation.
\begin{table}[h]
\centering
\caption{\textbf{Inference cost of the second pass's attention.} None: the \lrtok{} pass alone.
Cross-attention: \hrtok{} tokens read the cached \lrtok{} tokens only. One-way attention (MiSTR):
\hrtok{} tokens read the cached \lrtok{} tokens and each other. Self-attention: all \lrtok{} and
\hrtok{} tokens re-encoded jointly. PASTIS (64$\times$64 px, 12 months in 4 seasons), \lrtok{} =
8$\times$8 px seasonal tokens, \hrtok{} = 4$\times$4 px monthly tokens, 25\% of points selected
with one season each (212 selected point-months, 848 \hrtok{} tokens per sample). Same model and selections;
GMACs per sample, mean over 256 test samples, including the \lrtok{} pass. $^\dagger$Summed over both passes.}
\label{tab:oneway-cost}
\footnotesize
\begin{tabular}{l c c c}
\toprule
Attention & tokens / block$^\dagger$ & GMACs (attn.) & 2nd pass \\
\midrule
none                & 1024 & 10.4 (4.8)  & -- \\
cross-attention     & 1872 & 18.3 (8.8)  & 7.9 \\
one-way attention   & 1872 & \textbf{22.9} (12.7) & \textbf{12.5} \\
self-attention      & 2896 & 37.2 (21.6) & 26.8 \\
\bottomrule
\end{tabular}
\end{table}

On PASTIS with $S_8T_\text{sea}$ \& $S_4T_\text{mon}$ ($q = 1/4$), this makes the second pass 2.1$\times$ cheaper than self-attention, and inference 1.6$\times$ cheaper overall (\cref{tab:oneway-cost}).

\begin{figure}[t]
    \centering
    \includegraphics[width=\linewidth]{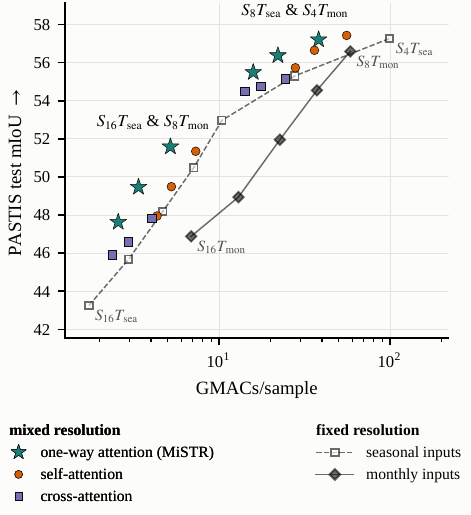}
    \caption{\textbf{One-way attention matches self-attention at lower cost.} PASTIS, trained with each second-pass attention design from the same guided warm-up; points are $q \in \{1/8, 1/4, 1/2\}$, means over seeds. Self-attention re-encodes the \lrtok{} tokens and costs more for the same accuracy; cross-attention is cheaper but loses accuracy, as the \hrtok{} tokens cannot read each other.}
    \label{fig:oneway-trained}
\end{figure}
\begin{table}[t]
\centering
\caption{\textbf{Trained second-pass attention designs} (\cref{fig:oneway-trained}). PASTIS; every model is trained from the same guided warm-up per configuration and seed and differs only in its second-pass attention. Pairs $S_c$ \& $S_f$ = $S_cT_\text{sea}$ \& $S_fT_\text{mon}$. GMACs per sample and test mIoU, mean $\pm$ standard error over three seeds.}
\label{tab:oneway-trained}
\scriptsize
\setlength{\tabcolsep}{2.3pt}
\begin{tabular}{l c rl rl rl}
\toprule
 & & \multicolumn{2}{c}{one-way (MiSTR)} & \multicolumn{2}{c}{self-attention} & \multicolumn{2}{c}{cross-attention} \\
\cmidrule(lr){3-4}\cmidrule(lr){5-6}\cmidrule(lr){7-8}
Pair & $q$ & GMACs & mIoU & GMACs & mIoU & GMACs & mIoU \\
\midrule
$S_{16}$ \& $S_{8}$ & 1/8 & 2.6 & 47.6$\pm$0.1 & 4.3 & 48.0$\pm$0.2 & 2.4 & 45.9$\pm$0.1 \\
 & 1/4 & 3.4 & 49.5$\pm$0.2 & 5.3 & 49.5$\pm$0.2 & 2.9 & 46.6$\pm$0.1 \\
 & 1/2 & 5.2 & 51.6$\pm$0.1 & 7.3 & 51.3$\pm$0.2 & 4.0 & 47.8$\pm$0.0 \\
\midrule
$S_{8}$ \& $S_{4}$ & 1/8 & 15.8 & 55.5$\pm$0.4 & 27.9 & 55.7$\pm$0.1 & 14.1 & 54.5$\pm$0.3 \\
 & 1/4 & 22.1 & 56.4$\pm$0.3 & 36.1 & 56.6$\pm$0.2 & 17.5 & 54.8$\pm$0.2 \\
 & 1/2 & 38.2 & 57.2$\pm$0.2 & 55.7 & 57.4$\pm$0.1 & 24.4 & 55.2$\pm$0.1 \\
\bottomrule
\end{tabular}
\end{table}

These models are trained separately from those of \cref{fig:result} and \cref{tab:selector}, so the one-way numbers differ by up to 0.3 mIoU from the same configurations there.
Trained with each design (\cref{fig:oneway-trained}, \cref{tab:oneway-trained}), one-way attention is within 0.4 mIoU of self-attention at 57--71\% of its computation, and at equal computation it is more accurate.
Cross-attention saves a further 8--36\% but loses 1.0--3.8 mIoU, and the loss grows with the budget.

\subsection{What the \hrtok{} tokens attend to}
\label{apx:one-way-analysis}
\begin{table}[t]
\centering
\caption{\textbf{What MiSTR's \hrtok{} tokens attend to}: mean attention (\%) that \hrtok{} queries give each key group (over all blocks and heads), mean $\pm$ standard error over three seeds. ``This month'' \hrtok{} = the \hrtok{} tokens of the query's own point and of the other selected points in the same month. ``Same loc.'' = the same spatial point in other months, and the \lrtok{} cells under it. Pairs $S_c$ \& $S_f$ = $S_cT_\text{sea}$ \& $S_fT_\text{mon}$.}
\label{tab:oneway-attn}
\footnotesize
\setlength{\tabcolsep}{4pt}
\resizebox{\columnwidth}{!}{%
\begin{tabular}{l ccccc}
\toprule
 & \multicolumn{2}{c}{this month / season} & \multicolumn{2}{c}{other months / seasons} & same \\
\cmidrule(lr){2-3}\cmidrule(lr){4-5}
MiSTR config & \hrtok{} & \lrtok{} & \hrtok{} & \lrtok{} & loc. \\
\midrule
PASTIS $S_{16}$ \& $S_8$      & 23.1\,{\scriptsize$\pm$0.2} & 15.7\,{\scriptsize$\pm$0.0} & 26.7\,{\scriptsize$\pm$0.2} & 34.5\,{\scriptsize$\pm$0.2} & 23.4\,{\scriptsize$\pm$0.0} \\
PASTIS $S_8$ \& $S_4$       & 20.0\,{\scriptsize$\pm$1.5} & 18.8\,{\scriptsize$\pm$0.7} & 28.7\,{\scriptsize$\pm$1.1} & 32.6\,{\scriptsize$\pm$0.3} & 13.5\,{\scriptsize$\pm$0.2} \\
PLANTEUR $S_{16}$ \& $S_8$    & 19.2\,{\scriptsize$\pm$0.7} & 15.0\,{\scriptsize$\pm$0.1} & 39.5\,{\scriptsize$\pm$0.4} & 26.3\,{\scriptsize$\pm$0.2} & 19.7\,{\scriptsize$\pm$0.1} \\
BioMassters $S_{16}$ \& $S_8$ & 26.8\,{\scriptsize$\pm$0.1} & 12.4\,{\scriptsize$\pm$0.1} & 34.2\,{\scriptsize$\pm$0.1} & 26.6\,{\scriptsize$\pm$0.0} & 21.3\,{\scriptsize$\pm$0.1} \\
YieldSAT $S_{16}$ \& $S_8$    & 26.0\,{\scriptsize$\pm$0.5} & 13.4\,{\scriptsize$\pm$0.4} & 37.1\,{\scriptsize$\pm$0.1} & 23.5\,{\scriptsize$\pm$0.8} & 14.6\,{\scriptsize$\pm$0.4} \\
\bottomrule
\end{tabular}}
\end{table}

To verify that \hrtok{} tokens not only receive context through one-way attention but also learn to use it, we study how they spend their attention in MiSTR's second, mixed-resolution pass.
\hrtok{} tokens give 37--51\% of their attention to \lrtok{} tokens and the rest to other \hrtok{} tokens, and they concentrate on their own location: the other modalities and band groups of the same spatiotemporal point, the same point at other timesteps, and the \lrtok{} cell it overlaps receive 13--23\% of the attention (\cref{tab:oneway-attn}).
MiSTR thus uses the spatiotemporal context available at all of its resolutions.

\begin{table}[t]
\centering
\caption{\textbf{Which context the trained model relies on}: change in the test metric when a key group is masked out of the \hrtok{} tokens' attention at every block, at test time only, for a model trained with all tokens (mIoU points$\uparrow$ on PASTIS and PLANTEUR, RMSE$\downarrow$ in t/ha on BioMassters and YieldSAT), mean $\pm$ standard error over three seeds. Pairs $S_c$ \& $S_f$ = $S_cT_\text{sea}$ \& $S_fT_\text{mon}$.}
\label{tab:oneway-mask}
\footnotesize
\setlength{\tabcolsep}{4pt}
\begin{tabular}{l ccc}
\toprule
 & \multicolumn{3}{c}{masked out} \\
\cmidrule(lr){2-4}
 & all & other & all but \\
MiSTR config & \lrtok{} & \hrtok{} & own spatial point \\
\midrule
PASTIS $S_{16}$ \& $S_8$      & $-$4.4\,{\scriptsize$\pm$0.1} & $-$3.7\,{\scriptsize$\pm$0.1} & $-$13.9\,{\scriptsize$\pm$0.4} \\
PASTIS $S_8$ \& $S_4$       & $-$4.9\,{\scriptsize$\pm$0.2} & $-$2.1\,{\scriptsize$\pm$0.0} & $-$9.5\,{\scriptsize$\pm$1.4} \\
PLANTEUR $S_{16}$ \& $S_8$    & $-$1.3\,{\scriptsize$\pm$0.3} & $-$2.4\,{\scriptsize$\pm$0.3} & $-$9.9\,{\scriptsize$\pm$1.3} \\
BioMassters $S_{16}$ \& $S_8$ & +.29\,{\scriptsize$\pm$.04} & +.77\,{\scriptsize$\pm$.01} & +4.17\,{\scriptsize$\pm$.06} \\
YieldSAT $S_{16}$ \& $S_8$    & +.097\,{\scriptsize$\pm$.021} & +.052\,{\scriptsize$\pm$.015} & +.645\,{\scriptsize$\pm$.177} \\
\bottomrule
\end{tabular}
\end{table}

To see which context the trained model relies on, we mask each key group out of the \hrtok{} tokens' attention at test time (\cref{tab:oneway-mask}), a probe of what the trained model uses.
Masking the \lrtok{} tokens hurts on every run, and so does masking the \hrtok{} tokens of the other selected points.
Restricting an \hrtok{} token to the \hrtok{} tokens of its own point hurts most.
With so few \hrtok{} tokens selected, a point on its own sees little of its surroundings, and one-way attention is what supplies that context to MiSTR's second pass.

\subsection{How the decoder combines \lrtok{} and \hrtok{} predictions}
\label{apx:one-way-decoder}
\begin{table}[t]
\centering
\caption{\textbf{How the decoder combines \lrtok{} and \hrtok{} tokens}, at pixels covered by \hrtok{} tokens. With one-way attention the \lrtok{} tokens are unchanged by the \hrtok{} tokens, so the output there is exactly $(1-w)\,p_{\lrtok{}} + w\,p_{\hrtok{}}$, where $p_{\lrtok{}}$ and $p_{\hrtok{}}$ are the pooled predictions of the \lrtok{} and \hrtok{} tokens and $w$ is the \hrtok{} tokens' total weight in the decoder's pool. Mean $\pm$ standard error over three seeds ($w$ percentiles: means). The combined prediction beats both inputs on every run. Regression RMSE is in normalized target units. Pairs $S_c$ \& $S_f$ = $S_cT_\text{sea}$ \& $S_fT_\text{mon}$.}
\label{tab:oneway-decoder}
\footnotesize
\setlength{\tabcolsep}{4pt}
\resizebox{\columnwidth}{!}{%
\begin{tabular}{l c ccc}
\toprule
 & \hrtok{} weight & \multicolumn{3}{c}{acc.\ (\%)$\uparrow$ / RMSE$\downarrow$} \\
\cmidrule(lr){3-5}
MiSTR config & mean (p10--p90) & \lrtok{} & \hrtok{} & final \\
\midrule
PASTIS $S_{16}$ \& $S_8$ & .74\,{\scriptsize$\pm$.01} (.56--.89) & 51.7\,{\scriptsize$\pm$0.0} & 58.5\,{\scriptsize$\pm$0.2} & \textbf{59.7\,{\scriptsize$\pm$0.2}} \\
PASTIS $S_8$ \& $S_4$ & .74\,{\scriptsize$\pm$.02} (.57--.89) & 54.7\,{\scriptsize$\pm$0.1} & 58.5\,{\scriptsize$\pm$0.2} & \textbf{59.5\,{\scriptsize$\pm$0.1}} \\
PLANTEUR $S_{16}$ \& $S_8$ & .59\,{\scriptsize$\pm$.01} (.39--.77) & 69.8\,{\scriptsize$\pm$0.3} & 68.2\,{\scriptsize$\pm$0.3} & \textbf{71.3\,{\scriptsize$\pm$0.5}} \\
BioMassters $S_{16}$ \& $S_8$ & .57\,{\scriptsize$\pm$.01} (.45--.68) & .805\,{\scriptsize$\pm$.000} & .776\,{\scriptsize$\pm$.002} & \textbf{.731\,{\scriptsize$\pm$.001}} \\
YieldSAT $S_{16}$ \& $S_8$ & .61\,{\scriptsize$\pm$.05} (.37--.84) & .672\,{\scriptsize$\pm$.027} & .621\,{\scriptsize$\pm$.008} & \textbf{.599\,{\scriptsize$\pm$.006}} \\
\bottomrule
\end{tabular}}
\end{table}

Where \hrtok{} tokens are selected, the decoder gives them most of the weight ($w$ = 0.57--0.74) and lets them overrule the \lrtok{} prediction.
The combined prediction outperforms both the \lrtok{} and the \hrtok{} predictions alone on every run (\cref{tab:oneway-decoder}).
This holds even where the \hrtok{} tokens alone are weaker than the \lrtok{} ones.
On PLANTEUR, the \hrtok{} tokens alone reach 68.2\% accuracy against the \lrtok{} prediction's 69.8\%, yet the combination reaches 71.3\%.
The \lrtok{} and \hrtok{} predictions thus carry complementary information, and the temporal decoder uses both.
\FloatBarrier

\section{Initializing MiSTR with other RSFMs}
\label{apx:alt-rsfm-init}
MiSTR needs no new parameters, only a ViT that can tokenize patches of different sizes, so it can start from RSFMs other than OlmoEarth.
We initialize it with TerraMind-1.0-Tiny \citep{jakubik2025terramind}, a multimodal (Sentinel-2 and Sentinel-1) but single-time model: it was pre-trained on single scenes, has no temporal encoding, and processes a series one timestep at a time.
Many RSFMs share this design \citep{cong2022satmae,fuller2023croma,jakubik2025terramind}, and are applied to SITS by encoding each timestep on its own and aggregating over time \citep{marsocci2024pangaea}; we use TerraMind as a representative of this family to show that MiSTR adapts flexibly to such models.
A single-time model is also a harder test than OlmoEarth, as MiSTR can no longer trade temporal resolution against spatial resolution inside one sequence and can only allocate higher spatial resolution where it is needed.

\begin{figure}[t]
    \centering
    \includegraphics[width=\linewidth]{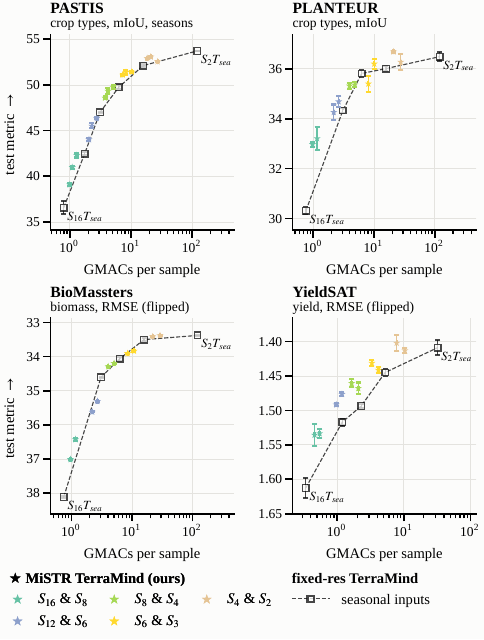}
    \caption{\textbf{Space-only MiSTR TerraMind.} \lrtok{} and \hrtok{} tokens both from seasonal composites, against fixed-resolution TerraMind-1.0-Tiny on the same composites (test metric vs.\ GMACs per SITS). Stars: MiSTR pairs at budgets $q \in \{1/8, 1/4, 1/2\}$ (PASTIS) or $\{1/4, 1/2\}$ (others). Error bars: $\pm 1$ standard error over three seeds.}
    \label{fig:terramind}
\end{figure}
\paragraph{TerraMind backbone.}
We run TerraMind's encoder on each seasonal composite separately, with the Sentinel-2 and Sentinel-1 tokens of a season in one sequence, and the temporal decoder (\cref{subsec:decoder}) pools across seasons.
As for OlmoEarth, patches of different sizes share TerraMind's patch embedding (PI-resized \citep{beyer2023flexivit} from its $16\times16$ patches), and each token's 2D sin-cos position encodes its patch center.
We add a learned, zero-initialized month encoding, so that the model is exactly the pre-trained TerraMind at the start of fine-tuning.

\paragraph{Space-only MiSTR.}
Every token, \lrtok{} and \hrtok{} alike, comes from the seasonal composites: MiSTR selects points by the uncertainty of the \lrtok{} prediction as before (\cref{subsec:selection}), and re-encodes each selected point at the finer patch size from the composite of the season its decoder attends to most, so that the temporal resolution stays fixed.
The \hrtok{} tokens of a point join the sequence of their season, with one-way attention to that season's \lrtok{} tokens (\cref{subsec:one-way-attention}).
We pair $S_pT_\text{sea}$ with $S_{p/2}T_\text{sea}$ for $p \in \{16, 12, 8, 6, 4\}$ at budgets $q \in \{1/8, 1/4, 1/2\}$ on PASTIS and $q \in \{1/4, 1/2\}$ elsewhere.
The baselines are fixed-resolution TerraMind models on the same seasonal composites, $S_{16}T_\text{sea}$ to $S_2T_\text{sea}$, with the same decoder; all models follow the training recipe of \cref{sec:results}.

\paragraph{Results.}
Across the four datasets, 40 of the 45 MiSTR TerraMind points lie above the fixed-resolution TerraMind curve (\cref{fig:terramind}).
For example, on PASTIS $S_{16}T_\text{sea}$ \& $S_8T_\text{sea}$ ($q = 1/2$) matches fixed $S_{12}T_\text{sea}$ with 26\% fewer GMACs, and on YieldSAT $S_4T_\text{sea}$ \& $S_2T_\text{sea}$ ($q = 1/4$) is more accurate than fixed $S_2T_\text{sea}$ with $4.1\times$ fewer GMACs.
On PASTIS and PLANTEUR the gains are largest at coarse resolutions and shrink as the coarse patches get finer, where less of the tile is left to resolve; on YieldSAT they stay large down to $S_4T_\text{sea}$ \& $S_2T_\text{sea}$.
MiSTR therefore applies beyond multi-temporal models like OlmoEarth: even a model with no notion of time can spend higher spatial resolution only where its \lrtok{} prediction is uncertain.

\begin{figure}[t]
    \centering
    \includegraphics[width=\linewidth]{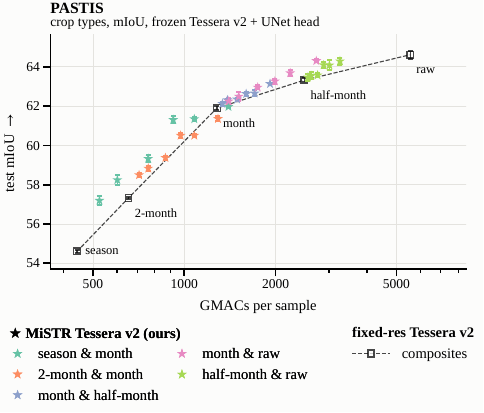}
    \caption{\textbf{Time-only MiSTR Tessera v2.} PASTIS, Tessera-v2-small with a UNet head. MiSTR re-encodes the most attended coarse points at a finer temporal resolution; stars of one color are different temporal budgets. Mean $\pm$ standard error over three training seeds of the head.}
    \label{fig:tessera}
\end{figure}
\paragraph{Time-only MiSTR Tessera v2.}
TerraMind tests MiSTR as a space-only selector; Tessera v2 \citep{feng2026tessera} tests the time-only case.
Tessera v2 encodes each pixel's time series independently and pools over time with attention, so selection is purely temporal.
We keep Tessera-v2-small frozen and train a UNet head on its pixel embeddings.
MiSTR encodes every pixel at a coarse temporal resolution (seasonal, 2-month, monthly or half-month composites) and re-encodes the windows that its attention pool, with queries tuned to the task, weights most at a finer one (monthly or half-month composites, or every acquisition).
The budget is the number of re-encoded windows per pixel, from 1/4 to 3; below one, only the pixels whose coarse prediction has the highest entropy re-encode a window.
27 of the 30 MiSTR points lie above the fixed-resolution curve (\cref{fig:tessera}).
For example, month \& raw at three windows per pixel reaches $64.3{\pm}0.1$ mIoU, within 0.3 mIoU of encoding every acquisition ($64.6{\pm}0.2$) at 49\% of its GMACs.
\FloatBarrier

\section{Matched Accuracy and Matched Compute}
\label{apx:matched-compute}
\providecolor{mbetter}{RGB}{0,128,128}
\providecolor{mworse}{RGB}{128,0,0}
\providecolor{mdom}{gray}{0.6}
\begin{table*}[t]
\centering
\caption{\textbf{Compute at matched accuracy.} For each fixed-resolution model, the cheapest MiSTR configuration that is at least as accurate (\textcolor{mbetter}{teal} $\Delta$), or, if no cheaper configuration is as accurate, the most accurate cheaper one (\textcolor{mworse}{maroon} $\Delta$). $\Delta$: MiSTR's metric gain over the fixed model (mIoU points; RMSE reduction), $\pm$ its standard error over seeds; saving: the fixed model's GMACs over MiSTR's. $^\dagger$, gray: off the fixed-resolution Pareto front, i.e.\ dominated by a cheaper, at least as accurate fixed-resolution model. Bold: the largest saving at matched accuracy against a Pareto-optimal fixed model. $S_{16}T_\text{sea}$ is omitted, as no MiSTR configuration is cheaper. Means over seeds of the points in \cref{fig:result}; GMACs per sample on the test split.}
\label{tab:matched-compute}
\scriptsize
\setlength{\tabcolsep}{3pt}
\begin{tabular}{l@{\hspace{8pt}}rrrr!{\color{black!25}\vrule}rrrr!{\color{black!25}\vrule}rrrr!{\color{black!25}\vrule}rrrr}
\toprule
 & \multicolumn{4}{c}{\textbf{PASTIS (mIoU)}} & \multicolumn{4}{c}{\textbf{PLANTEUR (mIoU)}} & \multicolumn{4}{c}{\textbf{BioMassters (RMSE $\downarrow$)}} & \multicolumn{4}{c}{\textbf{YieldSAT (RMSE $\downarrow$)}} \\
\cmidrule(lr){2-5}\cmidrule(lr){6-9}\cmidrule(lr){10-13}\cmidrule(lr){14-17}
 & \multicolumn{2}{c}{Fixed Res} & \multicolumn{2}{c}{MiSTR} & \multicolumn{2}{c}{Fixed Res} & \multicolumn{2}{c}{MiSTR} & \multicolumn{2}{c}{Fixed Res} & \multicolumn{2}{c}{MiSTR} & \multicolumn{2}{c}{Fixed Res} & \multicolumn{2}{c}{MiSTR} \\
\cmidrule(lr){2-3}\cmidrule(lr){4-5}\cmidrule(lr){6-7}\cmidrule(lr){8-9}\cmidrule(lr){10-11}\cmidrule(lr){12-13}\cmidrule(lr){14-15}\cmidrule(lr){16-17}
Model & metric & GMACs & $\Delta$ & saving & metric & GMACs & $\Delta$ & saving & metric & GMACs & $\Delta$ & saving & metric & GMACs & $\Delta$ & saving \\
\midrule
$S_{14}T_\text{sea}$ & 45.7 & 3.0 & \textcolor{mbetter}{+1.7{\tiny$\pm$0.2}} & \textcolor{mbetter}{1.1$\times$} & 33.0 & 2.9 & \textcolor{mbetter}{+0.6{\tiny$\pm$0.5}} & \textcolor{mbetter}{1.2$\times$} & 37.15 & 2.9 & \textcolor{mbetter}{+0.12{\tiny$\pm$0.05}} & \textcolor{mbetter}{1.1$\times$} & 1.440 & 1.5 & \textcolor{mworse}{$-$0.012{\tiny$\pm$0.010}} & \textcolor{mbetter}{1.1$\times$} \\
$S_{12}T_\text{sea}$ & 48.2 & 4.7 & \textcolor{mbetter}{+1.5{\tiny$\pm$0.2}} & \textcolor{mbetter}{1.4$\times$} & 34.4 & 4.6 & \textcolor{mbetter}{+0.8{\tiny$\pm$0.3}} & \textcolor{mbetter}{1.4$\times$} & 36.14 & 4.6 & \textcolor{mbetter}{+0.00{\tiny$\pm$0.02}} & \textcolor{mbetter}{1.0$\times$} & 1.425 & 2.4 & \textcolor{mbetter}{+0.024{\tiny$\pm$0.012}} & \textcolor{mbetter}{1.1$\times$} \\
$S_{10}T_\text{sea}$ & 50.5 & 7.1 & \textcolor{mbetter}{+1.1{\tiny$\pm$0.2}} & \textcolor{mbetter}{1.3$\times$} & 35.3 & 7.0 & \textcolor{mbetter}{+0.4{\tiny$\pm$0.2}} & \textcolor{mbetter}{1.2$\times$} & 35.23 & 7.0 & \textcolor{mworse}{$-$0.19{\tiny$\pm$0.13}} & \textcolor{mbetter}{1.4$\times$} & 1.417 & 3.5 & \textcolor{mbetter}{+0.016{\tiny$\pm$0.014}} & \textcolor{mbetter}{1.7$\times$} \\
$S_{8}T_\text{sea}$ & 53.0 & 10.4 & \textcolor{mbetter}{+0.4{\tiny$\pm$0.1}} & \textcolor{mbetter}{1.1$\times$} & 35.9 & 10.3 & \textcolor{mbetter}{+0.1{\tiny$\pm$0.2}} & \textcolor{mbetter}{1.5$\times$} & 34.23 & 10.3 & \textcolor{mworse}{$-$0.58{\tiny$\pm$0.02}} & \textcolor{mbetter}{1.0$\times$} & 1.414 & 5.0 & \textcolor{mbetter}{+0.013{\tiny$\pm$0.014}} & \textcolor{mbetter}{2.4$\times$} \\
$S_{6}T_\text{sea}$ & 55.3 & 27.7 & \textcolor{mbetter}{+0.4{\tiny$\pm$0.2}} & \textcolor{mbetter}{1.8$\times$} & 37.2 & 27.5 & \textcolor{mbetter}{+0.3{\tiny$\pm$0.2}} & \textcolor{mbetter}{1.8$\times$} & 33.48 & 27.6 & \textcolor{mworse}{$-$0.02{\tiny$\pm$0.04}} & \textcolor{mbetter}{1.2$\times$} & \textcolor{mdom}{1.382$^\dagger$} & \textcolor{mdom}{12.3} & \textcolor{mdom}{+0.010{\tiny$\pm$0.010}} & \textcolor{mdom}{3.4$\times$} \\
$S_{4}T_\text{sea}$ & 57.3 & 99.3 & \textcolor{mbetter}{+0.4{\tiny$\pm$0.4}} & \textcolor{mbetter}{\textbf{2.2$\times$}} & 38.0 & 98.7 & \textcolor{mbetter}{+0.0{\tiny$\pm$0.5}} & \textcolor{mbetter}{2.3$\times$} & 32.51 & 99.2 & \textcolor{mworse}{$-$0.41{\tiny$\pm$0.05}} & \textcolor{mbetter}{1.5$\times$} & 1.357 & 40.6 & \textcolor{mbetter}{+0.005{\tiny$\pm$0.006}} & \textcolor{mbetter}{\textbf{4.6$\times$}} \\
\midrule
$S_{16}T_\text{mon}$ & \textcolor{mdom}{46.9$^\dagger$} & \textcolor{mdom}{6.9} & \textcolor{mdom}{+0.5{\tiny$\pm$0.2}} & \textcolor{mdom}{2.6$\times$} & \textcolor{mdom}{32.4$^\dagger$} & \textcolor{mdom}{6.9} & \textcolor{mdom}{+1.1{\tiny$\pm$0.7}} & \textcolor{mdom}{2.7$\times$} & \textcolor{mdom}{36.80$^\dagger$} & \textcolor{mdom}{6.2} & \textcolor{mdom}{+0.49{\tiny$\pm$0.03}} & \textcolor{mdom}{1.8$\times$} & 1.436 & 2.4 & \textcolor{mbetter}{+0.035{\tiny$\pm$0.011}} & \textcolor{mbetter}{1.1$\times$} \\
$S_{14}T_\text{mon}$ & \textcolor{mdom}{48.9$^\dagger$} & \textcolor{mdom}{13.0} & \textcolor{mdom}{+0.8{\tiny$\pm$0.0}} & \textcolor{mdom}{3.8$\times$} & \textcolor{mdom}{34.1$^\dagger$} & \textcolor{mdom}{13.1} & \textcolor{mdom}{+1.1{\tiny$\pm$0.3}} & \textcolor{mdom}{3.9$\times$} & \textcolor{mdom}{35.86$^\dagger$} & \textcolor{mdom}{11.7} & \textcolor{mdom}{+0.44{\tiny$\pm$0.04}} & \textcolor{mdom}{2.3$\times$} & 1.416 & 4.1 & \textcolor{mbetter}{+0.015{\tiny$\pm$0.011}} & \textcolor{mbetter}{2.0$\times$} \\
$S_{12}T_\text{mon}$ & \textcolor{mdom}{52.0$^\dagger$} & \textcolor{mdom}{22.7} & \textcolor{mdom}{+0.3{\tiny$\pm$0.2}} & \textcolor{mdom}{3.1$\times$} & 36.0 & 22.9 & \textcolor{mbetter}{+0.5{\tiny$\pm$0.2}} & \textcolor{mbetter}{\textbf{2.4$\times$}} & \textcolor{mdom}{35.01$^\dagger$} & \textcolor{mdom}{20.5} & \textcolor{mdom}{+0.19{\tiny$\pm$0.04}} & \textcolor{mdom}{2.1$\times$} & 1.399 & 6.8 & \textcolor{mbetter}{+0.009{\tiny$\pm$0.013}} & \textcolor{mbetter}{2.2$\times$} \\
$S_{10}T_\text{mon}$ & \textcolor{mdom}{54.5$^\dagger$} & \textcolor{mdom}{37.3} & \textcolor{mdom}{+1.1{\tiny$\pm$0.4}} & \textcolor{mdom}{2.5$\times$} & \textcolor{mdom}{36.2$^\dagger$} & \textcolor{mdom}{37.9} & \textcolor{mdom}{+0.4{\tiny$\pm$0.3}} & \textcolor{mdom}{3.9$\times$} & \textcolor{mdom}{34.24$^\dagger$} & \textcolor{mdom}{33.7} & \textcolor{mdom}{+0.10{\tiny$\pm$0.06}} & \textcolor{mdom}{2.1$\times$} & 1.380 & 10.5 & \textcolor{mbetter}{+0.008{\tiny$\pm$0.011}} & \textcolor{mbetter}{2.9$\times$} \\
$S_{8}T_\text{mon}$ & 56.6 & 58.5 & \textcolor{mbetter}{+0.6{\tiny$\pm$0.2}} & \textcolor{mbetter}{1.5$\times$} & \textcolor{mdom}{36.4$^\dagger$} & \textcolor{mdom}{59.5} & \textcolor{mdom}{+0.2{\tiny$\pm$0.2}} & \textcolor{mdom}{6.2$\times$} & 33.20 & 52.8 & \textcolor{mbetter}{+0.12{\tiny$\pm$0.03}} & \textcolor{mbetter}{1.3$\times$} & 1.374 & 15.7 & \textcolor{mbetter}{+0.002{\tiny$\pm$0.007}} & \textcolor{mbetter}{4.3$\times$} \\
$S_{6}T_\text{mon}$ & 59.0 & 181.6 & \textcolor{mworse}{$-$0.1{\tiny$\pm$0.3}} & \textcolor{mbetter}{1.6$\times$} & \textcolor{mdom}{37.8$^\dagger$} & \textcolor{mdom}{185.0} & \textcolor{mdom}{+0.3{\tiny$\pm$0.1}} & \textcolor{mdom}{4.4$\times$} & 32.47 & 163.1 & \textcolor{mbetter}{+0.06{\tiny$\pm$0.11}} & \textcolor{mbetter}{\textbf{1.3$\times$}} & 1.345 & 43.8 & \textcolor{mbetter}{+0.003{\tiny$\pm$0.005}} & \textcolor{mbetter}{2.1$\times$} \\
\bottomrule
\end{tabular}
\end{table*}

\providecolor{mbetter}{RGB}{0,128,128}
\providecolor{mworse}{RGB}{128,0,0}
\providecolor{mdom}{gray}{0.6}
\begin{table*}[t]
\centering
\caption{\textbf{Accuracy at matched compute.} For each fixed-resolution model, the most accurate MiSTR configuration that needs no more GMACs; $\Delta$ is \textcolor{mbetter}{teal} where MiSTR is better and \textcolor{mworse}{maroon} where it is worse. $\Delta$: MiSTR's metric gain over the fixed model (mIoU points; RMSE reduction), $\pm$ its standard error over seeds; saving: the fixed model's GMACs over MiSTR's. $^\dagger$, gray: off the fixed-resolution Pareto front, i.e.\ dominated by a cheaper, at least as accurate fixed-resolution model. Bold: the largest gain against a Pareto-optimal fixed model. $S_{16}T_\text{sea}$ is omitted, as no MiSTR configuration is cheaper. Means over seeds of the points in \cref{fig:result}; GMACs per sample on the test split.}
\label{tab:matched-accuracy}
\scriptsize
\setlength{\tabcolsep}{3pt}
\begin{tabular}{l@{\hspace{8pt}}rrrr!{\color{black!25}\vrule}rrrr!{\color{black!25}\vrule}rrrr!{\color{black!25}\vrule}rrrr}
\toprule
 & \multicolumn{4}{c}{\textbf{PASTIS (mIoU)}} & \multicolumn{4}{c}{\textbf{PLANTEUR (mIoU)}} & \multicolumn{4}{c}{\textbf{BioMassters (RMSE $\downarrow$)}} & \multicolumn{4}{c}{\textbf{YieldSAT (RMSE $\downarrow$)}} \\
\cmidrule(lr){2-5}\cmidrule(lr){6-9}\cmidrule(lr){10-13}\cmidrule(lr){14-17}
 & \multicolumn{2}{c}{Fixed Res} & \multicolumn{2}{c}{MiSTR} & \multicolumn{2}{c}{Fixed Res} & \multicolumn{2}{c}{MiSTR} & \multicolumn{2}{c}{Fixed Res} & \multicolumn{2}{c}{MiSTR} & \multicolumn{2}{c}{Fixed Res} & \multicolumn{2}{c}{MiSTR} \\
\cmidrule(lr){2-3}\cmidrule(lr){4-5}\cmidrule(lr){6-7}\cmidrule(lr){8-9}\cmidrule(lr){10-11}\cmidrule(lr){12-13}\cmidrule(lr){14-15}\cmidrule(lr){16-17}
Model & metric & GMACs & $\Delta$ & saving & metric & GMACs & $\Delta$ & saving & metric & GMACs & $\Delta$ & saving & metric & GMACs & $\Delta$ & saving \\
\midrule
$S_{14}T_\text{sea}$ & 45.7 & 3.0 & \textcolor{mbetter}{\textbf{+1.7}{\tiny$\pm$0.2}} & \textcolor{mbetter}{1.1$\times$} & 33.0 & 2.9 & \textcolor{mbetter}{+0.6{\tiny$\pm$0.5}} & \textcolor{mbetter}{1.2$\times$} & 37.15 & 2.9 & \textcolor{mbetter}{\textbf{+0.12}{\tiny$\pm$0.05}} & \textcolor{mbetter}{1.1$\times$} & 1.440 & 1.5 & \textcolor{mworse}{$-$0.012{\tiny$\pm$0.010}} & \textcolor{mbetter}{1.1$\times$} \\
$S_{12}T_\text{sea}$ & 48.2 & 4.7 & \textcolor{mbetter}{+1.5{\tiny$\pm$0.2}} & \textcolor{mbetter}{1.4$\times$} & 34.4 & 4.6 & \textcolor{mbetter}{+0.8{\tiny$\pm$0.3}} & \textcolor{mbetter}{1.4$\times$} & 36.14 & 4.6 & \textcolor{mbetter}{+0.00{\tiny$\pm$0.02}} & \textcolor{mbetter}{1.0$\times$} & 1.425 & 2.4 & \textcolor{mbetter}{+0.024{\tiny$\pm$0.012}} & \textcolor{mbetter}{1.1$\times$} \\
$S_{10}T_\text{sea}$ & 50.5 & 7.1 & \textcolor{mbetter}{+1.1{\tiny$\pm$0.2}} & \textcolor{mbetter}{1.3$\times$} & 35.3 & 7.0 & \textcolor{mbetter}{+0.7{\tiny$\pm$0.1}} & \textcolor{mbetter}{1.0$\times$} & 35.23 & 7.0 & \textcolor{mworse}{$-$0.19{\tiny$\pm$0.13}} & \textcolor{mbetter}{1.4$\times$} & 1.417 & 3.5 & \textcolor{mbetter}{+0.027{\tiny$\pm$0.015}} & \textcolor{mbetter}{1.1$\times$} \\
$S_{8}T_\text{sea}$ & 53.0 & 10.4 & \textcolor{mbetter}{+0.4{\tiny$\pm$0.1}} & \textcolor{mbetter}{1.1$\times$} & 35.9 & 10.3 & \textcolor{mbetter}{+0.7{\tiny$\pm$0.2}} & \textcolor{mbetter}{1.1$\times$} & 34.23 & 10.3 & \textcolor{mworse}{$-$0.58{\tiny$\pm$0.02}} & \textcolor{mbetter}{1.0$\times$} & 1.414 & 5.0 & \textcolor{mbetter}{+0.042{\tiny$\pm$0.012}} & \textcolor{mbetter}{1.4$\times$} \\
$S_{6}T_\text{sea}$ & 55.3 & 27.7 & \textcolor{mbetter}{+0.8{\tiny$\pm$0.2}} & \textcolor{mbetter}{1.2$\times$} & 37.2 & 27.5 & \textcolor{mbetter}{+0.3{\tiny$\pm$0.2}} & \textcolor{mbetter}{1.8$\times$} & 33.48 & 27.6 & \textcolor{mworse}{$-$0.02{\tiny$\pm$0.04}} & \textcolor{mbetter}{1.2$\times$} & \textcolor{mdom}{1.382$^\dagger$} & \textcolor{mdom}{12.3} & \textcolor{mdom}{+0.030{\tiny$\pm$0.008}} & \textcolor{mdom}{1.4$\times$} \\
$S_{4}T_\text{sea}$ & 57.3 & 99.3 & \textcolor{mbetter}{+1.0{\tiny$\pm$0.3}} & \textcolor{mbetter}{1.6$\times$} & 38.0 & 98.7 & \textcolor{mbetter}{+0.2{\tiny$\pm$0.5}} & \textcolor{mbetter}{1.6$\times$} & 32.51 & 99.2 & \textcolor{mworse}{$-$0.41{\tiny$\pm$0.05}} & \textcolor{mbetter}{1.5$\times$} & 1.357 & 40.6 & \textcolor{mbetter}{+0.022{\tiny$\pm$0.013}} & \textcolor{mbetter}{1.1$\times$} \\
\midrule
$S_{16}T_\text{mon}$ & \textcolor{mdom}{46.9$^\dagger$} & \textcolor{mdom}{6.9} & \textcolor{mdom}{+4.7{\tiny$\pm$0.2}} & \textcolor{mdom}{1.3$\times$} & \textcolor{mdom}{32.4$^\dagger$} & \textcolor{mdom}{6.9} & \textcolor{mdom}{+3.2{\tiny$\pm$0.5}} & \textcolor{mdom}{1.2$\times$} & \textcolor{mdom}{36.80$^\dagger$} & \textcolor{mdom}{6.2} & \textcolor{mdom}{+1.38{\tiny$\pm$0.03}} & \textcolor{mdom}{1.2$\times$} & 1.436 & 2.4 & \textcolor{mbetter}{+0.035{\tiny$\pm$0.011}} & \textcolor{mbetter}{1.1$\times$} \\
$S_{14}T_\text{mon}$ & \textcolor{mdom}{48.9$^\dagger$} & \textcolor{mdom}{13.0} & \textcolor{mdom}{+4.9{\tiny$\pm$0.2}} & \textcolor{mdom}{1.2$\times$} & \textcolor{mdom}{34.1$^\dagger$} & \textcolor{mdom}{13.1} & \textcolor{mdom}{+2.5{\tiny$\pm$0.3}} & \textcolor{mdom}{1.4$\times$} & \textcolor{mdom}{35.86$^\dagger$} & \textcolor{mdom}{11.7} & \textcolor{mdom}{+1.43{\tiny$\pm$0.12}} & \textcolor{mdom}{1.0$\times$} & 1.416 & 4.1 & \textcolor{mbetter}{\textbf{+0.043}{\tiny$\pm$0.007}} & \textcolor{mbetter}{1.1$\times$} \\
$S_{12}T_\text{mon}$ & \textcolor{mdom}{52.0$^\dagger$} & \textcolor{mdom}{22.7} & \textcolor{mdom}{+4.1{\tiny$\pm$0.2}} & \textcolor{mdom}{1.0$\times$} & 36.0 & 22.9 & \textcolor{mbetter}{\textbf{+1.4}{\tiny$\pm$0.1}} & \textcolor{mbetter}{1.5$\times$} & \textcolor{mdom}{35.01$^\dagger$} & \textcolor{mdom}{20.5} & \textcolor{mdom}{+1.12{\tiny$\pm$0.03}} & \textcolor{mdom}{1.3$\times$} & 1.399 & 6.8 & \textcolor{mbetter}{+0.031{\tiny$\pm$0.008}} & \textcolor{mbetter}{1.2$\times$} \\
$S_{10}T_\text{mon}$ & \textcolor{mdom}{54.5$^\dagger$} & \textcolor{mdom}{37.3} & \textcolor{mdom}{+1.5{\tiny$\pm$0.4}} & \textcolor{mdom}{1.7$\times$} & \textcolor{mdom}{36.2$^\dagger$} & \textcolor{mdom}{37.9} & \textcolor{mdom}{+1.3{\tiny$\pm$0.3}} & \textcolor{mdom}{2.4$\times$} & \textcolor{mdom}{34.24$^\dagger$} & \textcolor{mdom}{33.7} & \textcolor{mdom}{+0.74{\tiny$\pm$0.04}} & \textcolor{mdom}{1.4$\times$} & 1.380 & 10.5 & \textcolor{mbetter}{+0.028{\tiny$\pm$0.009}} & \textcolor{mbetter}{1.2$\times$} \\
$S_{8}T_\text{mon}$ & 56.6 & 58.5 & \textcolor{mbetter}{+1.1{\tiny$\pm$0.4}} & \textcolor{mbetter}{1.3$\times$} & \textcolor{mdom}{36.4$^\dagger$} & \textcolor{mdom}{59.5} & \textcolor{mdom}{+1.6{\tiny$\pm$0.2}} & \textcolor{mdom}{1.4$\times$} & 33.20 & 52.8 & \textcolor{mbetter}{+0.12{\tiny$\pm$0.03}} & \textcolor{mbetter}{1.3$\times$} & 1.374 & 15.7 & \textcolor{mbetter}{+0.028{\tiny$\pm$0.010}} & \textcolor{mbetter}{1.1$\times$} \\
$S_{6}T_\text{mon}$ & 59.0 & 181.6 & \textcolor{mworse}{$-$0.1{\tiny$\pm$0.3}} & \textcolor{mbetter}{1.6$\times$} & \textcolor{mdom}{37.8$^\dagger$} & \textcolor{mdom}{185.0} & \textcolor{mdom}{+0.4{\tiny$\pm$0.2}} & \textcolor{mdom}{2.9$\times$} & 32.47 & 163.1 & \textcolor{mbetter}{+0.06{\tiny$\pm$0.11}} & \textcolor{mbetter}{1.3$\times$} & 1.345 & 43.8 & \textcolor{mbetter}{+0.010{\tiny$\pm$0.013}} & \textcolor{mbetter}{1.2$\times$} \\
\bottomrule
\end{tabular}
\end{table*}

\Cref{fig:result} compares whole curves; here we read off two matched comparisons for each fixed-resolution model, the basis of the 1.3--4.6$\times$ of \cref{sec:intro,sec:results}.
We compare MiSTR against the \emph{Pareto front} of the fixed-resolution models, the strongest fixed-resolution baseline at every compute level: a fixed-resolution model that another fixed-resolution model beats at no more compute is off this front, marked as dominated ($^\dagger$, gray), and excluded from every claim below.
\Cref{tab:matched-compute} compares each fixed-resolution model with the cheapest MiSTR configuration that is at least as accurate, or, if none is, with the most accurate cheaper one.
Against the Pareto-optimal fixed-resolution models, the largest saving on each dataset, $2.2\times$ on PASTIS, $2.4\times$ on PLANTEUR, $1.3\times$ on BioMassters and $4.6\times$ on YieldSAT, gives the 1.3--4.6$\times$ of \cref{sec:intro}.
\Cref{tab:matched-accuracy} reverses the comparison: for each fixed-resolution model, the most accurate MiSTR configuration that needs no more GMACs.
MiSTR is more accurate than 28 of the 34 Pareto-optimal fixed-resolution models, by up to 1.7 mIoU on PASTIS, 1.4 mIoU on PLANTEUR, 0.12 RMSE on BioMassters and 0.043 RMSE on YieldSAT, and still uses fewer GMACs than the model it beats.

\section{Seed Variation of the Main Result}
\label{apx:seeds}
\begin{figure*}[t]
    \centering
    \includegraphics[width=\linewidth]{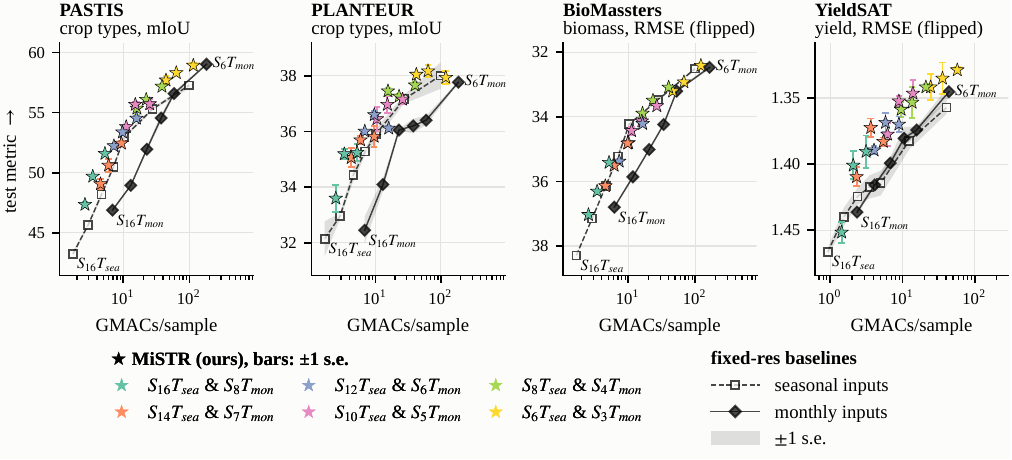}
    \caption{\textbf{The main result with its seed spread.} Means over three seeds as in \cref{fig:result}, with $\pm1$ standard error: shaded bands for the fixed-resolution curves, bars for MiSTR (horizontal bars where the number of selected points changes the cost across seeds). Where no bar or band is visible (e.g., PASTIS and BioMassters), the standard error is smaller than the marker.}
    \label{fig:result-std}
\end{figure*}

\Cref{fig:result} reports means over three seeds; \cref{fig:result-std} shows the same points with their standard error.
The standard errors are small compared with the gap between MiSTR and the fixed-resolution curves on PASTIS, PLANTEUR and YieldSAT.

\section{Flood Tracking on GEOID-Flood}
\label{apx:geoid-flood}
\Cref{sec:flood} trains flood tracking with temporal CutMix and leaves the backbone and training details to this appendix.
We give them here, then describe the training data and CutMix (\cref{apx:geoid-data}), predictions on a persistent and a short-lived flood (\cref{apx:geoid-samples}), and how the position of the labeled date in the series changes compute and accuracy (\cref{apx:geoid-position}).
All models use OlmoEarth-Tiny on $512\times512$-pixel tiles.
Fixed-resolution models train for 20 epochs.
Each MiSTR model starts from the fixed-resolution model at its coarse patch size and trains for 8 more epochs, drawing its selection budget per batch from 1, 2, 4 or 8\% of the series' high-resolution token slots, so an existing fixed-resolution model becomes a MiSTR model with a short fine-tuning.
\Cref{fig:geoid} shows one training seed (seed 0) of every model, with the fixed-resolution models at 20 epochs.
This appendix differs in both respects: its fixed-resolution models train for the same 28 epochs in total as MiSTR, and their scores are means over three seeds; MiSTR scores are seed 0, and the qualitative samples (\cref{fig:geoid-sample-full}) show seed 0 of every model.
The extra epochs do not explain MiSTR's gains: for $S_{32}$, the three-seed mean at 20 epochs is $67.6\pm1.1$ mIoU and $31.8\pm0.8$ flood bIoU (seed 0 alone, as in \cref{fig:geoid}: 68.0 and 31.8), and training for the same 8 more epochs changes it by $+0.5\pm0.6$ mIoU and $+1.7\pm1.0$ flood bIoU, to $68.1\pm1.1$ mIoU and $33.5\pm0.2$ flood bIoU (averaged over the event positions of \cref{fig:geoid}), while MiSTR $S_{64}$ \& $S_{16}$ at a budget of 8\% reaches 71.9 mIoU and 39.2 flood bIoU at less than a third of its GMACs.
Against these training-matched fixed-resolution models, the $7\times$ saving of \cref{fig:geoid} still holds: MiSTR $S_{64}$ \& $S_{16}$ at a budget of 2\% matches the 28-epoch $S_{32}$ (69.3 vs.\ $68.1\pm1.1$ mIoU, 33.5 vs.\ $33.5\pm0.2$ flood bIoU) with $7.2\times$ less compute (19.8 vs.\ 141.7 GMACs).
The gains therefore persist under a matched training budget: they come from MiSTR spending fine tokens where and when the flood appears.

\subsection{Training data and temporal CutMix}
\label{apx:geoid-data}
\begin{figure*}[t]
    \centering
    \includegraphics[width=\linewidth]{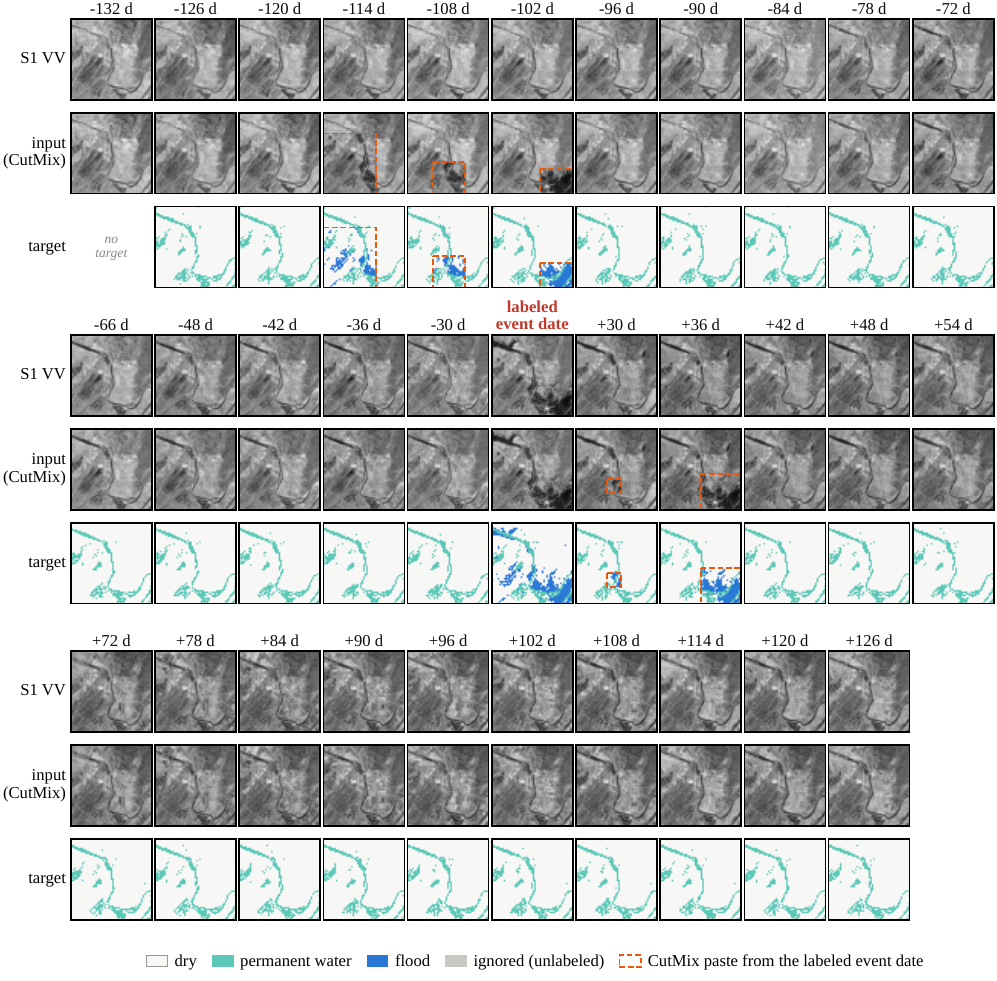}
    \caption{A 32-timestep training series of $512\times512$-pixel Sentinel-1 tiles augmented by temporal CutMix (timesteps in three rows of up to 11; per row: original S1 VV, CutMix input, training target). Flood regions of the GEOID-Flood labeled event date are pasted into 10--40\% of the other timesteps (dashed boxes), here timesteps before and after the labeled event date, and their targets become flood.}
    \label{fig:geoid-cutmix-sample}
\end{figure*}
For each event in GEOID-Flood \citep{Chiriaco2026GEOIDFlood}, we fetch a dense series of same-orbit Sentinel-1 acquisitions around the labeled date.
GEOID-Flood provides a flood map for one post-event image only, but tracking flood onset and recession needs a target at every timestep.
Unlike existing works on flood detection and segmentation benchmarks, which generally use curated pre- and post-event image pairs (as GEOID-Flood does), we aim to develop MiSTR to be computationally efficient under the realistic time-series setting where acquisitions come in sequentially and we do not know the date and location of new floods. We therefore propose a simple training recipe for tracking changes in SITS, using temporal CutMix to construct flood targets for the collected time series inputs.
Labeling every other timestep as flood-free would be wrong near the event: floods rarely come and go within one acquisition, and in many events the flood is already visible before the labeled date (onset) and lingers for weeks after it (recession), so this would result in incorrect supervision, with flooded pixels labeled as dry land.
We therefore drop all other acquisitions within 4 weeks of the labeled date from the training series.
The remaining timestep are then mostly clean negatives, with the permanent water of the GEOID-Flood label as its target and dry land elsewhere.
This creates a second hurdle, where each training series contains only one timestep with flood labels.
To provide flood targets away from the labeled date, we apply temporal CutMix \citep{yun2019cutmix}: the flood regions of the labeled date are partially pasted, together with their flood targets, into 10--40\% of the other timesteps, so that floods appear and persist at many points in the series.
The augmentation uses no annotation beyond the GEOID-Flood label.
\Cref{fig:geoid-cutmix-sample} shows a training series augmented by temporal CutMix.

Crucially, we make no modification to the evaluation series, so they keep every acquisition, and the event dates and labels are hidden from the model.
This represents a deployable setting, differing from existing flood detection setups: we do not require models to know when and where a flood happened; they predict continuously as acquisitions come in, to detect and segment new floods.
Scores are computed on the labeled date only, but the model is never told that only one timestep is scored, nor which one: it predicts every timestep of the series causally, and the scored prediction is one of them.
The labeled date is also not necessarily the first date on which the flood is visible: in many events the flood appears earlier (\cref{apx:geoid-samples}), so a model that does not track the flood while it persists scores poorly on the labeled date.

\subsection{Persistent and short-lived floods}
\label{apx:geoid-samples}
\begin{figure*}[p]
    \centering
    \includegraphics[width=1\linewidth]{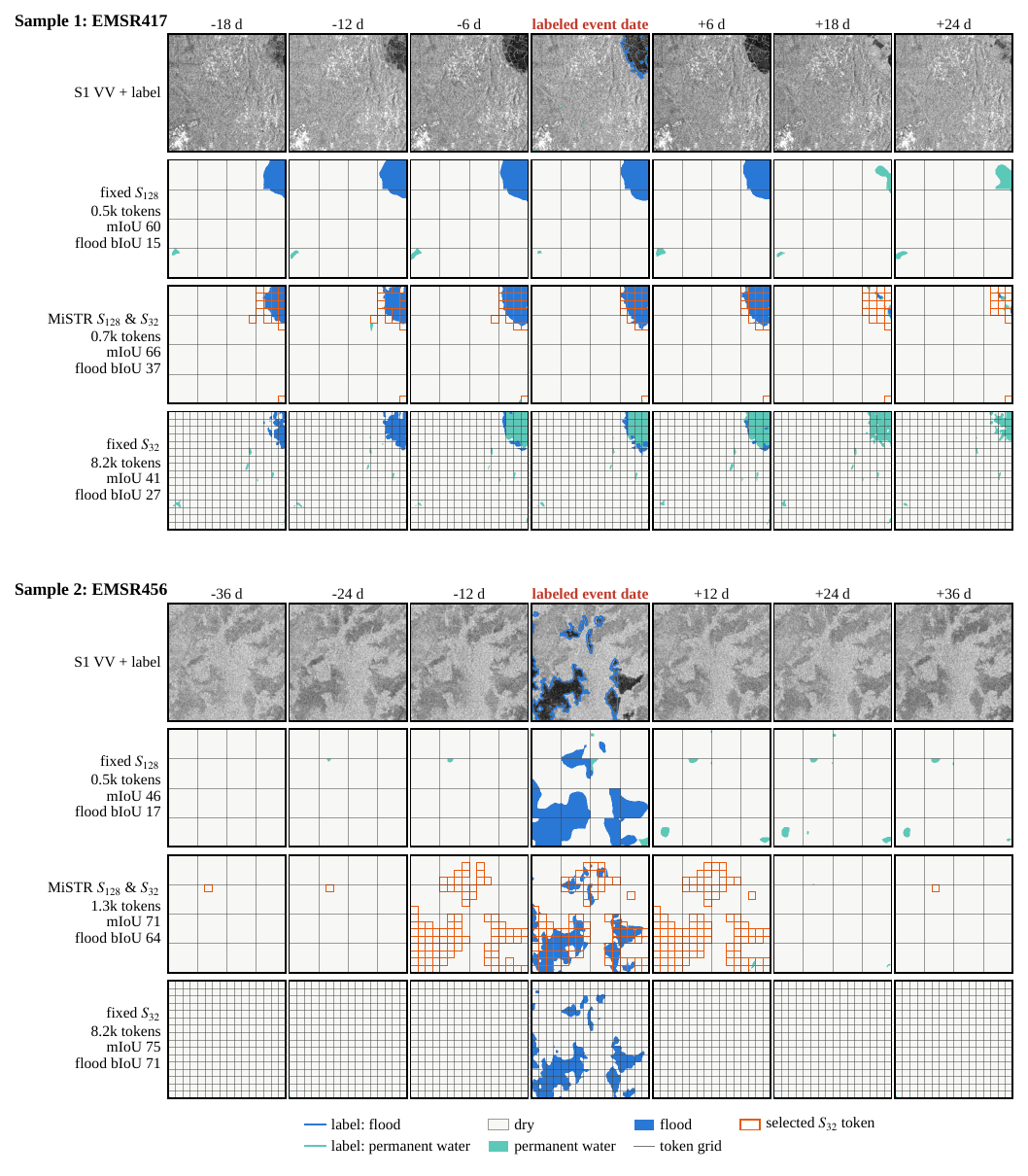}
    \caption{A persistent (sample 1) and a short-lived flood (sample 2), chosen from ten random test series with a flood ($512\times512$-pixel Sentinel-1 tiles; fixed-resolution models trained for as many epochs as MiSTR, \cref{apx:geoid-flood}). The GEOID-Flood label exists for one date only, drawn as outlines on S1 VV; every other timestep is predicted without supervision. Scores are on the labeled date.}
    \label{fig:geoid-sample-full}
\end{figure*}
\begin{figure*}[t]
    \centering
    \includegraphics[width=\linewidth]{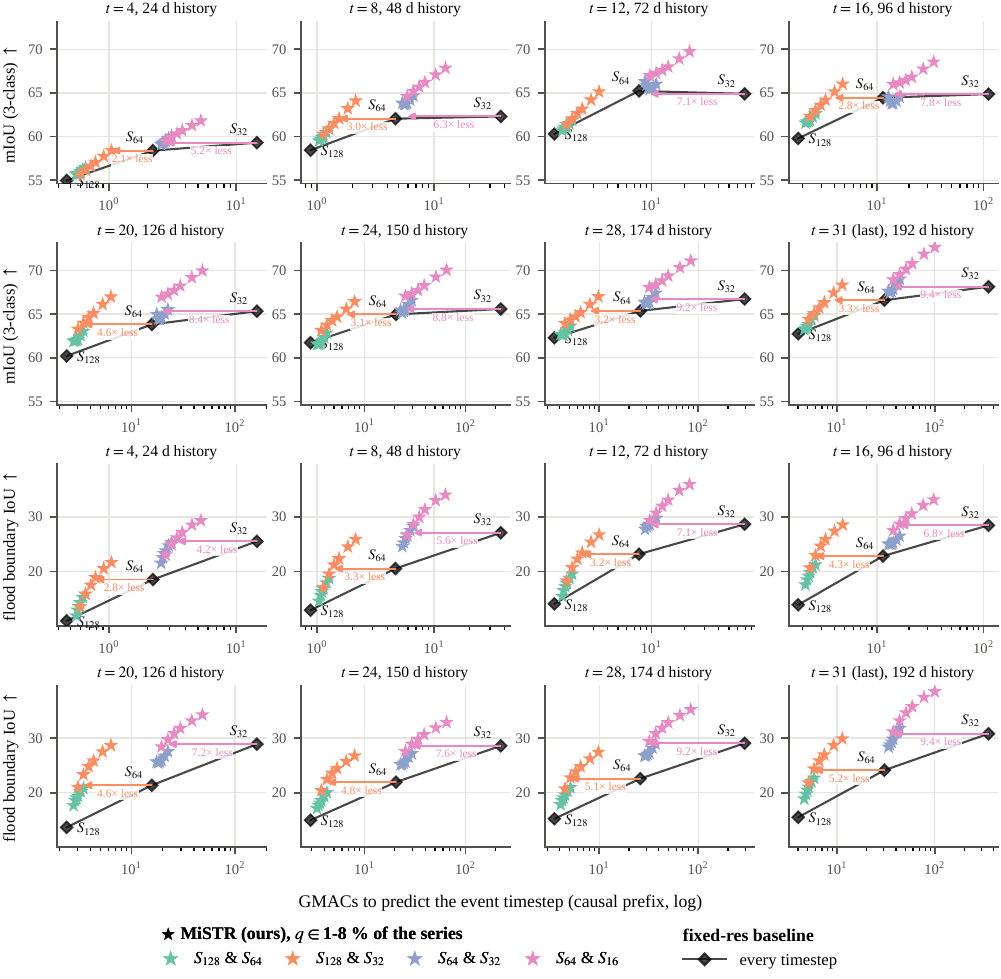}
    \caption{\textbf{MiSTR's saving and performance grow with history length.} Flood tracking on GEOID-Flood with the labeled date at timestep 4, 8, 12, 16, 20, 24, 28 or 31 (panels; 24 to 192 days of history), on the same test series in every panel. Rows 1--2: 3-class mIoU; rows 3--4: flood bIoU, against the GMACs needed to predict the labeled timestep (causal prefix, log scale). Stars: MiSTR at budgets of 1--8\% of the series' fine-token slots; diamonds: fixed resolution, mean over 3 seeds. Arrows mark the compute saving at matched performance.
    As the time series grow in length, MiSTR not only benefits computationally from only allocating high resolution tokens where and when they are needed, but also benefits in performance from the longer history context.}
    \label{fig:geoid-position}
\end{figure*}
To show how floods evolve in time, we drew ten test series at random among those with a flood (at least 1\% flood pixels on the labeled date) and chose two with common flood dynamics (\cref{fig:geoid-sample-full}).
In sample 1 the flood persists: it is visible in Sentinel-1 from 18 days before the labeled date to 6 days after it.
In sample 2 the flood is short-lived and visible on the labeled capture only.
Each sample shows seven timesteps around the labeled date (event at timestep 16 of 32).
Fixed $S_{128}$ (512 tokens) finds the flooded area but blurs its boundary (flood bIoU 15 and 17).
MiSTR $S_{128}$ \& $S_{32}$ (0.7k and 1.3k tokens) spends its $S_{32}$ tokens on the flood boundary, raising flood bIoU to 37 and 64.
In sample 1 it tracks the flood on every timestep where it is visible, while fixed $S_{32}$ (8.2k tokens) labels most of it as permanent water from 6 days before the labeled date on (flood bIoU 37 vs.\ 27).
In sample 2, MiSTR spends almost all of its high-resolution tokens on the flooded area at the timesteps around the labeled date, and largely recovers the predictions of fixed $S_{32}$ (flood bIoU 64 vs.\ 71) with \textbf{$6\times$ fewer tokens}.

\subsection{Event position}
\label{apx:geoid-position}
\Cref{fig:geoid} averages over the labeled date at timestep 12, 16, 20 and 24; \cref{fig:geoid-position} separates positions and extends the series to its end. To compare the effect of positions on the same events, it uses the 1{,}266 test series with 31 passes of history, each cut at the labeled date and placed at timestep 4, 8, 12, 16, 20, 24, 28 or 31; these series are harder than the full test set, so scores are lower than in \cref{fig:geoid}.
Because the models attend causally, a later event means a longer prefix to encode: moving the event from timestep 4 to timestep 31 raises the number of timesteps up to it from 5 to 32, and the cost of fixed $S_{32}$ grows 24-fold, from 14.5 to 354 GMACs, as attention grows superlinearly with the number of tokens.
The longer history also helps: from timestep 4 to 31, fixed $S_{32}$ gains $8.8\pm0.3$ mIoU and $5.3\pm0.9$ flood bIoU, mostly by recognizing permanent water, and MiSTR $S_{64}$ \& $S_{16}$ gains 10.8 and 9.2, most of its extra gain within the first eight timesteps.
With the fixed-resolution models trained for as many epochs as MiSTR, matching fixed $S_{32}$ costs $5.2\times$ to $9.4\times$ less compute in mIoU and $4.2\times$ to $9.4\times$ less in flood bIoU, rising with history.
At every position, MiSTR $S_{64}$ \& $S_{16}$ surpasses fixed $S_{32}$ on both metrics at a fraction of its compute.

\section{Temporal Datasets}
\label{apx:datasets}
\begin{figure*}[t]
    \centering
    \includegraphics[width=\linewidth]{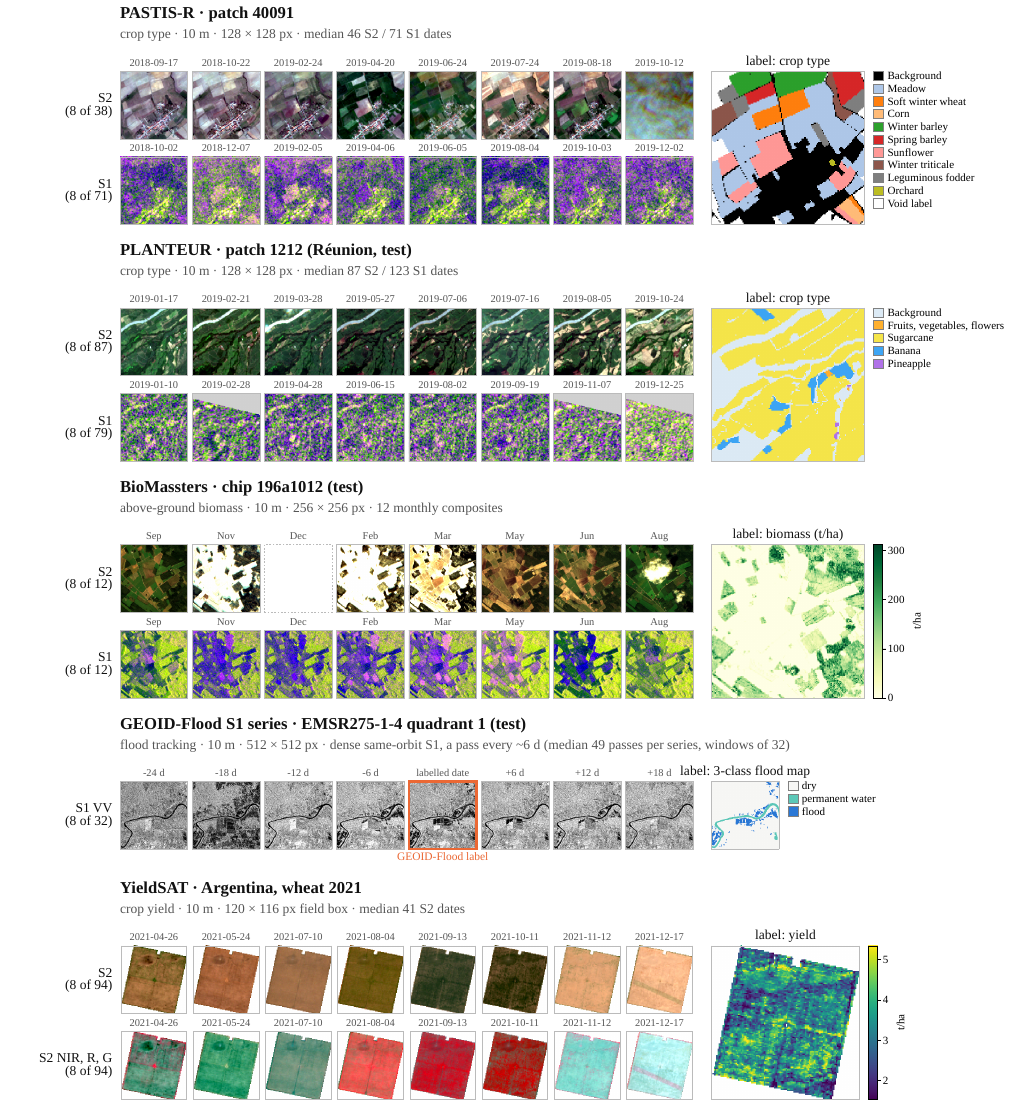}
    \caption{\textbf{One sample per dataset.} The five datasets cover per-pixel classification, per-pixel regression and change detection; each row shows a subset of the sample's timesteps next to its label.}
    \label{fig:datasets}
\end{figure*}

\Cref{tab:datasets} summarizes the five datasets of \cref{subsec:setup,sec:flood}, and \cref{fig:datasets} shows one sample of each.
They cover per-pixel classification (PASTIS, PLANTEUR), per-pixel regression (BioMassters, YieldSAT) and per-timestep change segmentation (GEOID-Flood).
For the four main datasets, models see $64\times64$-pixel tiles of each input; GEOID-Flood uses $512\times512$-pixel inputs.
Classification ignores void pixels, and regression is trained on standardized targets and evaluated in t/ha.

\begin{table}[h]
\centering
\caption{\textbf{Datasets.} Input band groups are OlmoEarth's (three for Sentinel-2, one for Sentinel-1); splits count patches (PASTIS, PLANTEUR), chips (BioMassters), fields (YieldSAT) or series (GEOID-Flood).}
\label{tab:datasets}
\scriptsize
\setlength{\tabcolsep}{2pt}
\begin{tabular}{@{}l l l l r@{}}
\toprule
Dataset & Task & Input & Target & Train / val / test \\
\midrule
PASTIS & segmentation & S2 + S1 (4) & 18 crops + bg. & 1{,}455 / 482 / 496 \\
PLANTEUR & segmentation & S2 + S1 (4) & 7 crops + bg. & 1{,}896 / 411 / 443 \\
BioMassters & regression & S2 + S1 (4) & biomass (t/ha) & 4{,}002 / 1{,}738 / 2{,}773 \\
YieldSAT & regression & S2 (3) & yield (t/ha) & 1{,}732 / 220 / 221 \\
GEOID-Flood & tracking & S1 (1) & dry / water / flood & 5{,}209 / 498 / 1{,}396 \\
\bottomrule
\end{tabular}
\end{table}

\paragraph{Series length.}
Each series is composited onto a fixed grid of calendar months, and every three months form a season.
PASTIS, PLANTEUR and BioMassters span one year (12 months, 4 seasons).
YieldSAT fields are imaged from about a month before seeding to a month after harvest, so each is placed on the 24 months of the two years ending with its harvest year (8 seasons); a sample covers 2--13 months (7 on average) and 1--5 seasons (3 on average).
Before compositing, the full series of a PASTIS sample has 38--61 Sentinel-2 and 65--71 Sentinel-1 acquisitions, a PLANTEUR sample 40--150 of each, and a YieldSAT sample 4--96 Sentinel-2 acquisitions; BioMassters provides one image per month, with Sentinel-2 missing in up to 7 of the 12 months.
Months without an acquisition produce no tokens: sequences are packed without padding, so neither MiSTR nor the fixed-resolution models pay for empty months, and the reported GMACs count only real tokens.

\paragraph{Code.} Code, configurations and model checkpoints to reproduce every experiment, figure and table in this paper are available at \url{https://github.com/timgzhou/MiSTR}.

\end{document}